\documentclass[lettersize,journal]{IEEEtran}
\usepackage{amsmath,amsfonts}
\usepackage{algorithmic}
\usepackage{algorithm}
\usepackage{array}
\usepackage[caption=false,font=normalsize,labelfont=sf,textfont=sf]{subfig}
\usepackage{textcomp}
\usepackage{stfloats}
\usepackage{url}
\usepackage{verbatim}
\usepackage{graphicx}
\usepackage{cite}
\usepackage[colorlinks=true,  % hyperref配置
citecolor=blue,
linkcolor=blue,
urlcolor=cyan]{hyperref}
\usepackage{multirow}
\usepackage{booktabs, multirow}
\usepackage[table]{xcolor}
\usepackage{graphicx}  % for \resizebox
\usepackage{makecell}  % for \Xhline
\usepackage{ulem}
\usepackage{arydshln}
\usepackage{tabularx}

\begin{document}

\title{ CADIC: Continual Anomaly Detection Based on Incremental Coreset}

\author{Gen Yang, Zhipeng Deng, Junfeng Man*
%	, Zhipeng Deng, Shuang Mei, Junfeng Man*
        % <-this % stops a space
\thanks{Gen Yang, Junfeng Man: Hunan First Normal University, Changsha, China.
}% <-this % stops a space
\thanks{Zhipeng Deng: Changsha University of Science and Technology, Changsha, China.}
%\thanks{Shuang Mei: China University of Geosciences, Wuhan, China.
%}% <-this % stops a space
\thanks{*Correspondence to: 751987807@qq.com}}

%\thanks{This paper was produced by the IEEE Publication Technology Group. They are in Piscataway, NJ.}% <-this % stops a space
%\thanks{Manuscript received April 19, 2021; revised August 16, 2021.}}
% The paper headers
%\markboth{Journal of \LaTeX\ Class Files,~Vol.~14, No.~8, August~2021}%
%{Shell \MakeLowercase{\textit{et al.}}: A Sample Article Using IEEEtran.cls for IEEE Journals}

%\IEEEpubid{0000--0000/00\$00.00~\copyright~2021 IEEE}
% Remember, if you use this you must call \IEEEpubidadjcol in the second
% column for its text to clear the IEEEpubid mark.

\maketitle

\begin{abstract}

The primary objective of Continual Anomaly Detection (CAD) is to learn the normal patterns of new tasks under dynamic data distribution assumptions while mitigating catastrophic forgetting. Existing embedding-based CAD approaches continuously update a memory bank with new embeddings to adapt to sequential tasks. However, these methods require constructing class-specific sub-memory banks for each task, which restricts their flexibility and scalability. To address this limitation, we propose a novel CAD framework where all tasks share a unified memory bank. During training, the method incrementally updates embeddings within a fixed-size coreset, enabling continuous knowledge acquisition from sequential tasks without task-specific memory fragmentation. In the inference phase, anomaly scores are computed via a nearest-neighbor matching mechanism, achieving state-of-the-art detection accuracy. We validate the method through comprehensive experiments on MVTec AD and Visa datasets. Results show that our approach outperforms existing baselines, achieving average image-level AUROC scores of 0.972 (MVTec AD) and 0.891 (Visa). Notably, on a real-world electronic paper dataset, it demonstrates 100\% accuracy in anomaly sample detection, confirming its robustness in practical scenarios. The implementation will be open-sourced on GitHub.

\end{abstract}


\section{Introduction}

\IEEEPARstart{A}{nomaly} Detection (AD) learns only anomaly-free samples and identify abnormal samples as instances that significantly differ from the learned behavior, playing a crucial role in industrial manufacturing. Most of the existing anomaly detection methods assume that the task is fixed. However, in real-world applications, new products could be added to the production line and there may exist concept drift for the same product during the producing process. These factors decrease the performance of anomaly detection models significantly. Continual Learning (CL) is a learning paradigm proposed to avoid Catastrophic Forgetting. It is also called lifelong learning, incremental learning or continuous learning. In the continual learning setting, the model can learn different tasks continuously. After learning new tasks, the model can still retain knowledge of past tasks. 
%Anomaly detection can identify anomalies, while continual learning can adapt models to new tasks. Researchers combine the two, resulting in Continual Anomaly Detection.

Continual Anomaly Detection (CAD) is a combination of AD and CL, which aims to learn new AD tasks while retaining performance on previously learned AD tasks. Most of current CAD methods are developed from reconstruction-based AD methods. By distilling previous AD models \cite{Yang} or retraining on newly updated dataset\cite{ref3}\cite{ref1}, the model can detect the anomalies of all the tasks based on the reconstruction error. Notably, embedding-based approaches \cite{SPADE}\cite{patchcore} represent another prominent category of AD methods. These approaches utilize frozen feature extractors to store features in a memory bank and detect anomalies based on the similarity between query embeddings and their nearest neighbors in the bank. Compared to reconstruction-based AD methods, embedding-based AD approaches are more stable due to the absence of neural network training processes, and they demonstrate superior anomaly detection performance by identifying subtle anomalies through local patch embeddings. Coincidentally, the replay-based continual learning uses a memory bank to store information of previous tasks. By replaying samples from the memory bank, it keeps the information of past tasks while learn new tasks. Thus, there is a gap between embedding based anomaly detection methods and replay-based continual learning methods, which motivate us to carry out this research.

%To fill the gap, few works have been conducted. Li et al. \cite{DNE} propose DNE that stores embeddings for each task in the memory bank during the training phase. At inference time, DNE reconstructs Gaussian distribution of each past task and use them to detect anomalous images. Liu et al. \cite{UCAD} propose UCAD that also stores all tasks' embeddings into memory bank and trains a prompt module to figure out the task types of the input images. In the inference stage, anomaly images are detected with the nearest-neighbor matching mechanism. Wu et al. proposed DFM method that integrates the feature extraction and matching modules into a single model. DFM trains adapters for each class and extract features of each class using a frozen backbone to form class-specific embeddings. During inference, the method distinguishes between different classes by retrieving the adapter corresponding to the nearest class-specific features. The above methods try to fill the gap between the embedding-based anomaly detection method and the continual learning method. However, they all need to form class-specific memory bank, which make them more near to traditional AD methods instead of CAD methods. 

To address the research gap at the intersection of embedding-based anomaly detection (AD) and continual learning (CL), several methodologies have been proposed. Li et al. \cite{DNE} introduce the DNE framework, which maintains a memory bank during training to store task-specific embeddings. During inference, Gaussian distributions of past tasks are reconstructed from this bank to identify anomalous inputs through statistical deviation scoring. Liu et al. \cite{UCAD} propose UCAD, which similarly stores multi-task embeddings in a memory bank but enhances it with a trainable prompt module to dynamically infer task types from input images. Anomaly detection is then performed via nearest-neighbor matching against task-specific prototypes. Wu et al. \cite{DFM} present the DFM method, which unifies embedding extraction and matching within a single model architecture. It employs class-specific adapter modules trained per category while using a frozen backbone for embedding extraction, forming class-conditional embeddings. During inference, the method distinguishes between different classes by retrieving the adapter corresponding to the nearest class-specific embeddings, and directly generates an anomaly map via the Feature Matching Network. While these approaches attempt to reconcile embedding-based AD with CL principles, they share a critical limitation: reliance on class-specific memory banks. This design choice aligns them more closely with conventional AD methods rather than achieving the task-agnostic and incremental learning objectives of CL, thereby leaving the core CL challenges of catastrophic forgetting and task-free adaptation unresolved.

\begin{figure*}[!t]
	\centering
	\includegraphics[width=0.7\textwidth]{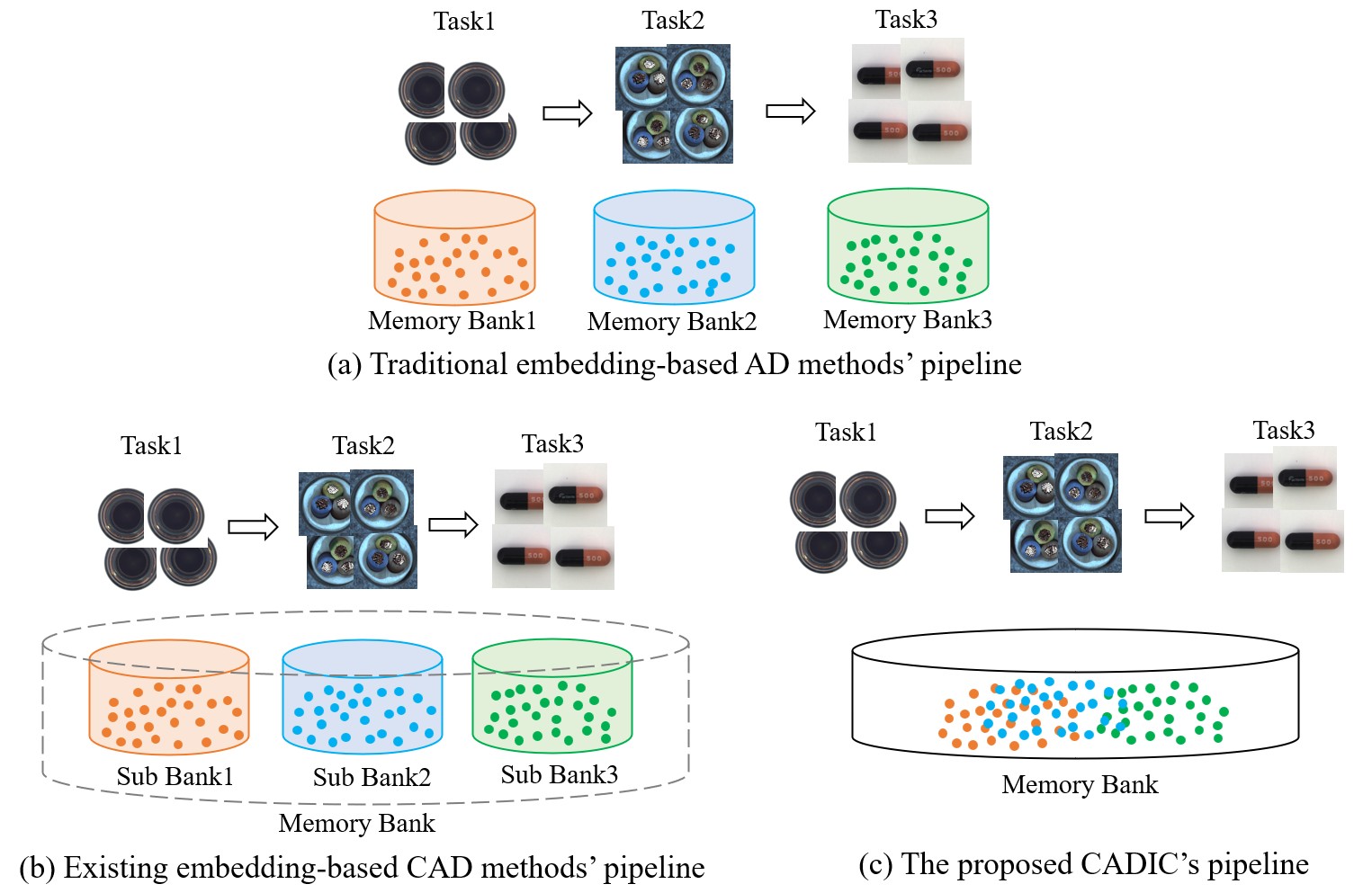}
	\caption{Comparison of AD, existing CAD and the proposed CADIC pipeline. (a) Traditional embedding-based AD methods use separate memory banks, each task has its own individual memory bank. (b) Existing embedding-based CAD methods imply only one memory bank, but each task has a specific sub bank. (c) Our embedding-based CAD method (CADIC) utilizes a single memory bank, and all tasks share it.}
	\label{pipeline}
\end{figure*}

Therefore, we introduce Continual Anomaly Detection Based on Incremental Coreset (CADIC) in this paper. As shown in Fig.\ref{pipeline}, traditional embedding-based AD methods \cite{SPADE}\cite{patchcore}\cite{padim} use separate memory banks, each task has its own individual memory bank. Existing embedding-based CAD methods \cite{DNE}\cite{UCAD} \cite{DFM} imply only one memory bank, but each task has a specific sub bank. On the contrary, the proposed CADIC utilizes a single memory bank, and all tasks share it. The main contributions are multifold as follows:
\begin{itemize}
	\item{we propose a novel embedding-based continual anomaly detection method. The method learn new tasks by updating the embeddings in the fixed-size memory bank incrementally, saving most typical embeddings of each task automatically.}
	\item{We perform continual anomaly detection experiments on the MVTec AD and VisA dataset. The results show that the proposed method achieves state-of-the-art performance.}
	\item{We verify the detection performance of the method for a real-scenario E-paper dataset. The results show that the proposed method can meet the requirements for industrial applications.}
\end{itemize}

%the embeddings of different tasks are put into the model sequentially. With the newly proposed data selecting method in this paper, embeddings in the fixed-size memory are updated incrementally. 
%to full fill the gap between anomaly detection and continual learning. our self-supervised dimension reduction block consists of a pollution module followed by a PCA(Principal component analysis) model. Noises are applied on the normal embeddings, hence producing many polluted embeddings. Even these embeddings are not from real abnormal samples, their distribution are quite different from normal embeddings' distribution. Next, the mixture of the normal embeddings and abnormal embeddings are used to calculate the PCA components. Then the dimensions of resulting embeddings are decreased by mapping on the PCA components. Compare to the previous MVG-based approach Padim, our proposed PcaAD exhibits superior detection quality. Moreover, PcaAD locates the various anomaly regions more accurately. 

\section{Related work}

\subsection{Anomaly Detection}
AD methods learn solely from anomaly-free samples but can detect various unknown anomalous samples. These methods are split into two main families: reconstruction-based methods and embedding-based methods.
%
% method only learn anomaly-free samples, and detect anomalies samples 
%Anomaly detection is an unsupervised learning method that only requires learning normal samples to detect anomalies and segment anomalous regions.%
%We can split most of the pixel-level anomaly detection approaches into two main families: reconstruction-based methods and feature embedding-based methods.%
%methods only learn anomaly-free samples, and detect anomalies samples 
%The goal of the image anomaly detection is to train a neural network using anomaly-free samples, and then, the network can detect anomalies and locate anomalous regions during inference.
%
% In the early years, anomaly detection was mainly based on traditional machine learning methods.
%In recent years, with the development of deep learning methods, more and more researchers have begun to deal with anomaly detection tasks with deep learning methods. Deep learning methods do not require manual feature designing and can be performed in an end-to-end manner. At present,
%anomaly detection based on deep learning methods can be mainly divided into two categories: anomaly detection methods based on representation estimation and anomaly detection methods based on reconstruction.

\subsubsection{Reconstruction-based Methods} 
The reconstruction-based methods learn to reconstruct normal images during training, and detect anomalies by evaluating the reconstruction errors in the inference stage. Typical methods like AnoGAN\cite{AnoGAN} and f-AnoGAN\cite{F-AnoGAN} reconstruct normal samples with generative adversarial networks. By comparing the difference between the generated samples the input samples, they detect anomalies at the pixel level. Unlike the methods above, which reconstruct normal samples, DRAEM \cite{DRAEM} reconstructs simulated anomaly samples. By training the model in an end-to-end manner, it directly outputs the anomaly maps of the input samples. Some other works employ autoencoders, vision transformers, distillation models, and diffusion models as reconstruction frameworks, and achieve competitive detection results.

\subsubsection{Embedding-based Methods}
The embedding-based methods use pre-trained neural networks to extract meaningful embeddings from images, and detect anomalies by analyzing the similarities between embeddings. Typical methods like SPADE\cite{SPADE} and PatchCore\cite{patchcore} save features in a memory bank, and calculate anomaly scores according to the distance between the image embeddings and their nearest neighbors in the memory bank. Works like PaDiM\cite{padim} and GAD\cite{GAD} establish multivariate Gaussian distributions for embeddings stored in the memory bank, and derive anomaly scores for image embeddings by evaluating their Mahalanobis Distances in the distribution. Other works like FastFlow\cite{fastflow} and CFLOW-AD\cite{Cflow} map embeddings into a standard normal distribution with the Normalizing Flow technique. Then, the anomaly scores for image embeddings are computed according to their probabilities in the distribution. In summary, the embedding-based anomaly detection methods use a memory bank to store the normal embeddings, and detect anomalies according to the similarities between embeddings and its nearest neighbors in the memory bank.

\subsection{Continual Learning}
In the continual learning setting, models can learn new tasks sequentially without forgetting knowledge of previous tasks. The implementations of continual learning can be divided into three families \cite{incremental_survey1} \cite{incremental_survey2}: parameter isolation methods, regularization methods, and memory replay methods. 
\subsubsection{Parameter isolation methods}
In this kind of method, specific network parameters are assigned to specific tasks. The parameters of networks are isolated in two main ways: mask out network parameters or add new networks for new tasks. For the former, PackNet\cite{PackNet} uses weight-based pruning techniques to generate a series of task-specific parameter masks. In the training stage,  only mask-free parameters are trained for new tasks. While in the inference stage, only masked parameters are used for prediction. For the latter, Expert Gate\cite{Expert Gate} uses multiple models to learn multiple tasks. In the training stage, a specialized model and expert gate are trained to learn each task. In the inference stage, the expert gate selects an appropriate model to deal with the task at hand.
\subsubsection{Regularization methods}
For regularization methods, the knowledge of previous tasks is preserved by introduce additional regularization terms. One implementation is to distill knowledge from a past model into the current model. LwF\cite{LwF} serves as a typical method of this implementation. When training on new tasks, LwF utilizes the output of the previous model as soft label for previous model, translating the knowledge of the former model to the next model, continuously. Another implementation is to penalize the changes of key neural network parameters. EWC\cite{EWC} is a representative of this method. It assesses the importance of parameters based on their posterior probabilities. When learning new tasks, only the less important parameters are encouraged to adapt to new tasks, while the important ones change little, preserving information from previous tasks effectively.
\subsubsection{Memory replay methods}
This line of works store samples from previous tasks into memory. When learning new tasks, samples in the memory are replayed to avoid forgetting. Samples are replayed in two main ways. First, samples in the memory are reused to train the model. Typically, iCaRL \cite{iCaRL} stores a subset of exemplars per class in the training stage, and classifies the testing data according to their distance to the means of all exemplars. Second, samples in the memory are reused to constrain the gradient direction of parameters. Typically, GEM \cite{GEM} calculates the loss gradient at the memory samples, and constrains the loss gradient at new tasks to the direction that does not decrease the former tasks' performance.
%Coincidentally, the replay-based continual learning uses a memory to store information of previous tasks incrementally. By replaying samples from the memory, it keeps the information of past tasks while learn new tasks. If we can update the embeddings in the memory incrementally, we can achieve anomaly detection in a continual learning manner.
%传统的异常检测方法为每个任务训练一个任务，不适用于持续任务的情形。
\subsection{Continual Anomaly Detection}
Traditional AD methods train separate models for each task, not fit for continuous tasks. CAD applies continual learning
 techniques into traditional AD methods, aiming to learn new anomaly detection tasks without forgetting old tasks. To lean anomaly detection tasks continuously, Yang et al.\cite{Yang} propose a reverse distillation-based method for this purpose. When learning a new task, it uses the distillation loss to learn new knowledge from the teacher model in this task. At the same time, it employs pooling distillation loss to learn old knowledge from the student model of the previous task. Pezze et al.\cite{ref1} introduce a reconstruction-based CAD approach for this purpose. They use a memory module to store old-task's images, and a detection module to reconstruct error for anomaly identification. ReplayCAD \cite{ReplayCAD} leverages compressed semantic embeddings and masks derived from previous tasks to guide a diffusion model in synthesizing samples from old tasks. IUF \cite{IUF} utilizes singular value decomposition to regulate gradient update directions by projecting gradients onto a subspace orthogonal to the feature representations of old tasks, though this process is computationally expensive. CDAD \cite{CDAD}  uses gradient projection to achieve stable continual learning, and it uses an iterative singular value decomposition method based on the transitive property of linear representation to improve computational inefficiency. CFRDC \cite{CFRDC} uses a context-aware feature reconstruction model to capture valuable anomaly identification-related knowledge, and utilize an intermediate feature organization strategy to avoid inter-class context conflict. DER \cite{DER} dynamically incorporates task-specific adapters and employs learnable prompts to query the corresponding adapter for each task, resulting in successful continual anomaly detection.
%In the CAD paradigm, the model is required to learn new normal knowledge from new tasks while preserving knowledge from old tasks. DNE \cite{DNE} alleviates forgetting by storing the mean and standard deviation of each task. However, this method is unable to achieve pixel-level anomaly localization. UCAD \cite{UCAD} overcomes this limitation by storing path-level image embeddings for each task. Moreover, they encode task-specific knowledge into the learnable prompts. Most recently, ReplayCAD \cite{ReplayCAD} employs compressed semantic embeddings and masks from previous tasks to guide a diffusion model in generating samples from old tasks. IUF \cite{IUF} uses singular value decomposition (SVD) to control the direction of gradient updates. It projects the gradient onto a subspace orthogonal to the feature representation of old tasks. However, it consumes a lot of time. CDAD \cite{CDAD} proposes iterative SVD to address this issue. In addition, DER \cite{DER} dynamically adds an adapter for each task and designs a learnable prompt to query the corresponding adapter.
\section{method}

\begin{figure*}[!t]
	\centering
	\includegraphics[width=0.7\textwidth]{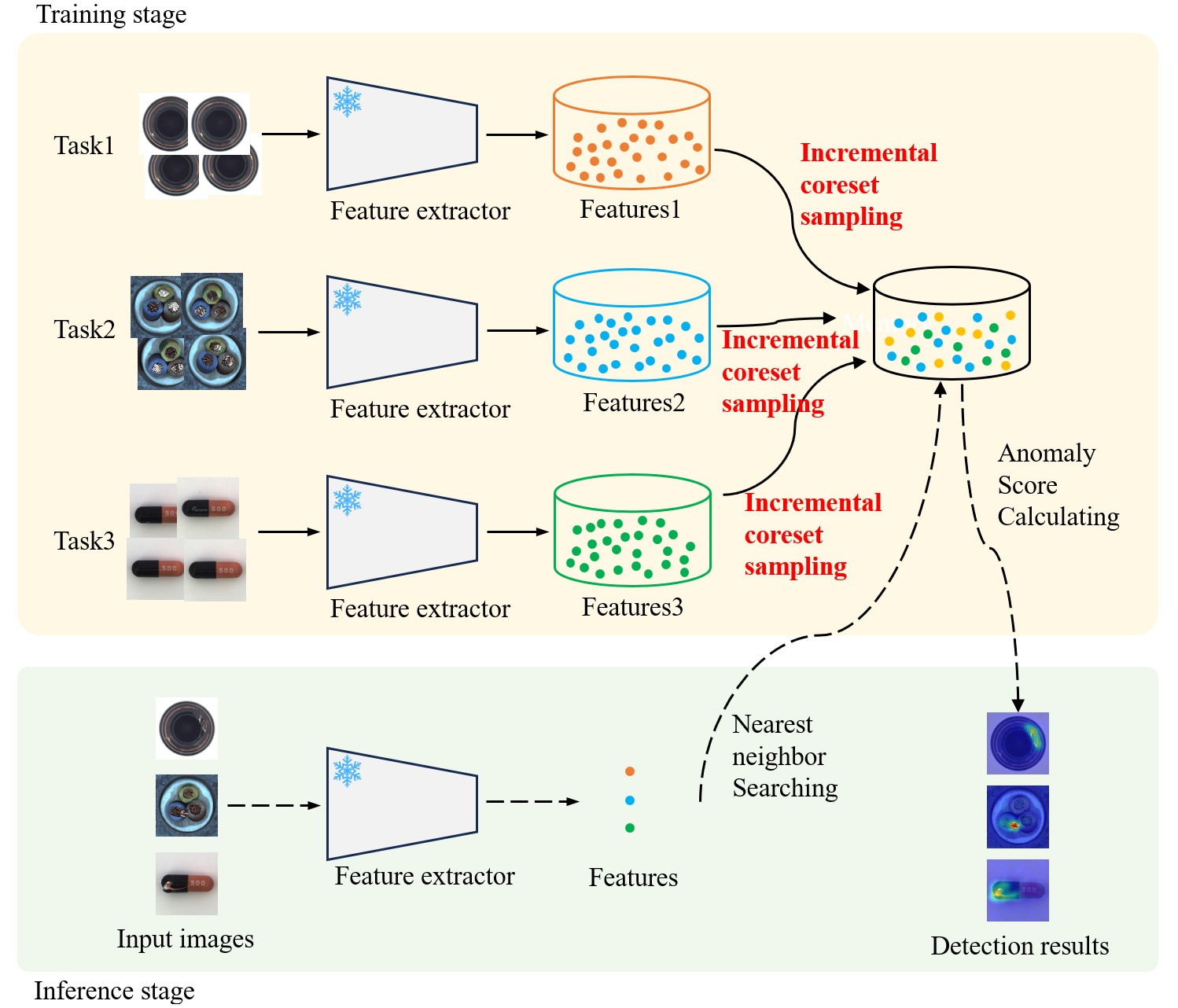}
	\caption{Framework for the proposed method.}
	\label{fig_1}
\end{figure*}

\subsection{Problem Definition}

In the  CAD paradigm, a unified model is trained non-repetitively on incrementally added tasks and tested on all tasks. To formulate this problem, we assume there are a total of $K$ sequential anomaly detection tasks $\left\{T_k|k\in (1,2,..., K)\right\}$. Each task $T_k$ corresponds to a dataset $D_k$. The train set includes only normal samples, while the test set contains both normal and abnormal samples. In each dataset, $X_k \in \mathbb{N}^{H\times W \times 3}$ stands for an image with height $H$ and width $W$, $Y_k\subset \left\{0, 1\right\}^{H\times W}$ indicates whether an image (or a pixel in the image) is normal $(0)$ or abnormal $(1)$. In the training stage, only the $k$-th task's training samples are used to train model. In the testing stage, test samples of all the previous tasks $\left\{T_k|k\in (1,2,..., K-1)\right\}$ are used to evaluate the model.
%the $k$ anomaly detection tasks are sent to the model one by one. During the inference stage, the test samples contain not only test samples of the current task $T_k$ but also test samples of all the previous tasks $\left\{T_n|n\in (1,2,...,k-1)\right\}$. The test samples of each task include both anomaly-free and anomalous samples.

%The goal is to train the model to learn a mapping $f_\theta: X \rightarrow \left\{0,1\right\}^{H\times W}$ from the space of images $X$ to a probability space, assigning a probability to each image (or each pixel in the image), indicating how likely it is to be normal or abnormal.

% the test samples of not only  the $k$th task but also all the previous tasks ${Tn|n\in {1,2,3,...,k-1}}$ are sent to the model for prediction. For every task, the test samples contain both  anomaly-free samples and anomalous samples. 

%Only anomaly-free samples $X$ are used to train the model.

\subsection{Feature extraction}
Relevant studies have demonstrated that analyzing the differences between embeddings extracted by pre-trained models can achieve highly accutate anomaly detction. Compared to pre-trained convolutional neural networks (CNNs)\cite{vgg}\cite{resnet}, pre-trained Vision Transformers (ViTs)\cite{vit} excel at capturing global dependencies between arbitrary locations within an image. When employing pre-trained models for anomaly detection, the following two aspects must be considered. First, deeper feature extraction layers. Shallow layers primarily encode low-level texture information, while deeper layers specialize in high-level semantic information. Since normal and abnormal discrimination heavily relies on semantic differences, deeper features provide more accurate anomaly descriptions. Second, larger spatial dimensions of feature maps. Larger feature maps enable finer-grained anomaly maps, resulting in more precise localization at the pixel level and more accurate detection at the image level.

Thus, we choose the \textit{ViT-Base-Patch8-224} model pre-trained on the ImageNet 21k as the feature extractor. The extractor contains $12$ transformer layers. For an input image $I \in \mathbb{R}^{224 \times 224 \times 3}$, each layer outputs a feature map $X\in \mathbb{R}^{28\times28\times768}$ that contains $28\times28$ embeddings  $x\in \mathbb{R}^{768}$, and each embedding represents a $8\times 8$ patch on the input image $I$. In this paper, we choose the $9$-th layer. By computing anomaly scores for each $x$, and upsampling the resulting matrix to the spatial dimensions of $I$, an anomaly map is generated. This anomaly map enables both image-level anomaly detection and pixel-level anomaly localization. Since the local structure of a high-dimensional manifold is homeomorphic to Euclidean space, we compute the anomaly score of an embedding by measuring the distance between the embedding and its nearest neighbors stored in a memory bank.
%and each embedding $x\in \mathbb{R}^{768}$ on $X$ correspond to a $8\times 8$ patch on $I$. 

%There are $28\times28$ embeddings on $X$, each embedding $x\in \mathbb{R}^{768}$ correspond to a $8\times 8$ image patch on $I$.

%For each embedding $x$ in on $X$, the feature map, it stands for a $8\times 8$ image patch on the input image $I$
%each layer outputs feature map with size $28\times28\times768$. For a specific layer, the obtained feature map is expressed as $X\in \mathbb{R}^{28 \times 28 \times768}$.

%We choose only one layer as extractor, then the obtained feature map is expressed as $X\in \mathbb{R}^{28 \times 28 \times768}$.
% For an input image $I \in \mathbb{R}^{224 \times 224 \times 3}$, the model obtain 12 feature maps whose size are all $28\times28\times768$. We choose only one layer as our extractor.
%the feature map size of each layer is $28\times28\times768$, i.e., $X\in \mathbb{R}^{28 \times 28 \times768}$.

\subsection{training stage}
%embeddings from normal image of the same category are always similar, but the quantity is large. Thus, a crucial work is to eliminate the redundant embeddings. The k-center method is proved to be an effective method to select 
%{Applying CL to unsupervised AD faces two challenges: 1)How to determine the task identities of the incoming image automatically; 2) How to guide the model’s predictions for the relevant task in an unsupervised manner. Thus, a continual prompting module is designed to dynamically adapt and instruct unsupervised model predictions.}

The greatest challenge that applying continual learning to unsupervised anomaly detection is how to add new tasks' anomaly-free knowledge into the anomaly detection model. To tackle this problem, we propose an incremental coreset updating mechanism, aiming  to store typical image embeddings across all tasks into embdeddiong coreset. Inspired by the mini-batch training paradigm, we update the memory bank with new task embeddings through dynamic batch processing. Specifically, when updating memory buffer $C$ with the current batch embeddings $X$, we firstly compute the maximum distance from each embedding in $X$ to all embeddings in $C$ as follows:
\begin{equation}
	\label{eqn_ex1}
	d_{max}(\mathbf{X},\mathbf{C}) = \max_{\mathbf{x}\in \mathbf{X}}\left\lbrace  d(\mathbf{x},\mathbf{C}) \right\rbrace.
\end{equation}
Then identify the corresponding specific feature element $x$ in $X$ that produces this maximum distance as follows:
\begin{equation}
	\label{eqn_ex2}
	\mathbf{x} = \mathop{\arg\max}_{\mathbf{x}\in \mathbf{X}}\left\lbrace  d(\mathbf{x},\mathbf{C}) \right\rbrace,
\end{equation}
where $d(x,C)$ is the distance between $x$ and its nearest neighbors in $C$ expressed as follows:
\begin{equation}
	\label{eqn_d}
	d(\mathbf{x},\mathbf{C})= \min_{\mathbf{c}\in \mathbf{C}} ||x-c||_2.
\end{equation}
Furthermore, we compute the minimum nearest neighbor distance between elements in memory bank $C$:
\begin{equation}
	\label{eqn_ex3}
	\lvert\mathbf{C}\rvert_{\min} = \min_{\mathbf{c_1},\mathbf{c_2}\in \mathbf{C}}\left\lbrace  d(\mathbf{c_1},\mathbf{c_2}) \right\rbrace,
\end{equation}
 and identify the corresponding element pair $(c_1, c_2)$:
\begin{equation}
	\label{eqn_ex4}
	\mathbf{c_1},\mathbf{c_2} =
	\mathop{\arg\min}_{\mathbf{c_1},\mathbf{c_2}\in \mathbf{C}}\left\lbrace  d(\mathbf{c_1},\mathbf{c_2}) \right\rbrace.
\end{equation}
When the condition
\begin{equation}
	\label{eqn_ex5}
	d_{max}\left( {\mathbf{X},\mathbf{C}} \right) \textgreater  \lvert\mathbf{C}\rvert_{\min}
\end{equation}
is satisfied, we add $x$ from $X$ into $C$, and delete $c_1$ from $C$. Subsequently, we iteratively execute the procedures defined in Equations (\ref{eqn_ex1})-(\ref{eqn_ex4}) until the condition (\ref{eqn_ex5}) is not satisfied, thereby completing the memory bank updating process. This replacement mechanism is universally applied to all tasks' images.

Directly computing the pairwise distances between every vector in $X$ and $C$ is computationally expensive. In this study, the distances between elements in $X$ and $C$ are efficiently calculated using matrix operations, with the formula defined as follows:
\begin{align}
	\label{eqn_ex6}
	D=&\nonumber sqrt\{[diag(XX^T)]^T\times \mathbf{1}_{1\times m}+\mathbf{1}_{n\times 1}\times diag(CC^T)\\& -2XC^T \},
\end{align}
where $diag(\cdot)$ converts the main diagonal elements of a matrix into a row vector while preserving their original order, $sqrt(\cdot)$ computes the arithmetic square root for each element in a matrix independently, $\mathbf{1}_{1\times m}$ is a row vector where all elements are 1, and $\mathbf{1}_{n\times 1}$ is a column vector where all elements are 1.

\begin{figure}[!t]
	\centering
	\includegraphics[width=0.4\textwidth]{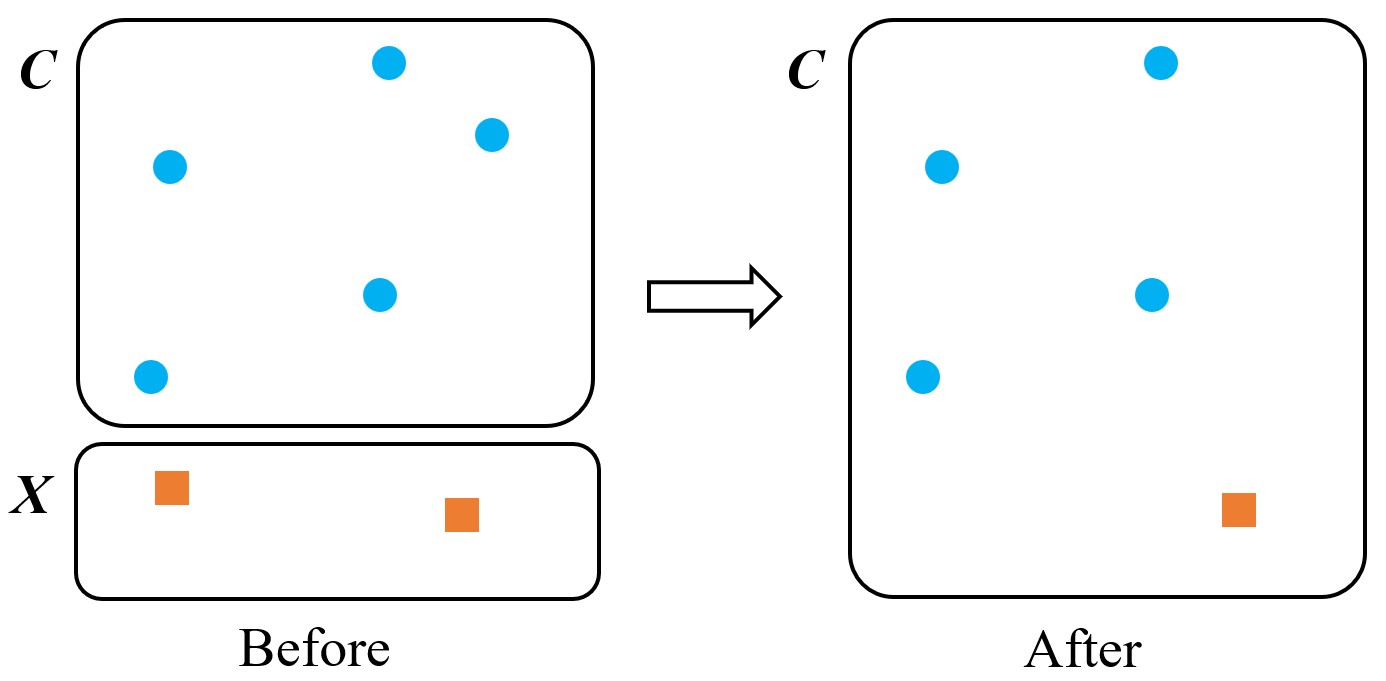}
	\caption{Illustration of the incremental coreset sampling.}
	\label{fig_3}
\end{figure}

\subsection{Inference Stage}
Since the memory bank only stores normal embeddings, anomaly scores of embeddings can be calculated by measuring their distance to their nearest neighbors in the coreset. A smaller distance indicates greater similarity to normal embeddings, leading to a lower anomaly score; a larger distance suggests less similarity to normal embeddings, resulting in a higher anomaly score. Following PatchCore\cite{patchcore}, we calculate the embedding anomaly score as follows:
\begin{equation}
	\label{eqn_pixel}
	s_{pix}= \min_{\mathbf{c}\in \mathbf{C}} ||x-c||_2.
\end{equation}
and calculate the image-level anomaly score as follows:
\begin{equation}
	\label{eqn_image}
	s_{img}=\left(1-\frac{\exp||x-c^*||_2}{\sum_{c\in \mathcal{N}_b(c^*)}\exp||x-c||_2}\right)\cdot s_{pix}^*,
\end{equation}
where $s_{pix}^*$ denotes the largest anomaly score among all embeddings of the test image, $c^*$ is the corresponding embeddings in the memory bank $C$, and $\mathcal{N}_b(c^*)$ stands for the $b$ embeddings in $C$ that are closest to $c^*$. 

%For image-level anomaly detection, the input image is predicted to be abnormal if $s_{img}$ exceeds a predefined image-level threshold, else it is normal. For pixel-level anomaly localization, the anomaly scores corresponding to the feature map are upsampled to the size of the input image, followed by smoothing and binarization, thereby achieving pixel-level anomaly localization.

For image-level anomaly detection, the input image is classified as anomalous if its anomaly score $s_{img}$ exceeds a predefined threshold; otherwise, it is deemed normal. For pixel-level anomaly localization, anomaly scores obtained from feature embeddings are spatially upsampled to align with the dimensions of the input image, after which anomalous regions are identified through binarization.

%the matrix composed of embedded anomaly values is first upsampled to the size of the input image, followed by spatial smoothing to generate the anomaly map. This map is then binarized using a specified pixel-level threshold, where the resulting foreground regions represent anomalous areas, thereby achieving pixel-level anomaly localization.

%The memory bank stores the features of normal images for various tasks, and the normality or abnormality of a feature can be determined based on its KNN distance to the features in the memory bank. Specifically, this is achieved by calculating the KNN distance between the feature in question and the normal features. If the KNN distance is greater than or equal to a set threshold, the feature is predicted as normal; if the KNN distance is less than the set density threshold, the feature is predicted as abnormal. In this way, normal and abnormal features are distinguished based on the features stored in the memory bank.

%
%\begin{equation}
%	\label{eqn_ex7}
%	s(p) = \min_{m\in M}||p-m||_2
%%	\mathop{\arg\max}_{m^{\text{text}}\in \mathbf{P}} \arg\min_{m\in K^t_n}
%\end{equation}
%
%\begin{equation}
%	\label{eqn_ex7}
%	S'(I) = \max_{p\in g(I)} s(p)
%\end{equation}
%
%\begin{equation}
%	\label{eqn_ex7}
%	S(I) = \left( 1-\frac{exp||m^{\text{text},*}-m^*||_2}{\sum_{m\in N_b(m*)}exp||m^{\text{text},*}-m||_2} \right)S'(I)
%\end{equation}
%

%\begin{equation}
%	\label{eqn_ex7}
%	m^{\text{text},*},m^* =
%	\mathop{\arg\max}_{m^{\text{text}}\in \mathbf{P}} \arg\min_{m\in K^t_n}
%%	\left\lbrace  d(\mathbf{c_1},\mathbf{c_2}) \right\rbrace.
%\end{equation}

\section{experiment}
The proposed method is evaluated on two widely-used anomaly detection datasets:
the MVTech AD\cite{MVTech} and the VisA\cite{VisA} . The MVTec AD dataset contains 15 subsets, and the VisA dataset contains 12 subsets. Each subset corresponds to a category of object. For each task, the training set contains only anomaly-free samples, while the testing set includes both anomaly-free samples and anomalous samples. We treat each subset as an independent anomaly detection task and order them alphabetically. Thus, there are 15 tasks on the MVTec dataset and 12 tasks on the VisA dataset. The model sequentially learns each task. After completing all tasks, we evaluate the detection metrics across all tasks.

All experiments are conducted on a workstation with 
an NVIDIA GeForce RTX 4070 GPU, an Intel i5-13600KF CPU, a 32GB of RAM. The Operating System is Windows 10.

%英特纳雄耐尔:
%GPU:Nvidia geforce 4070 CPU:intel i5~13600KF
%
%英特纳雄耐尔:
%Python:3.8.0 cuda11.8
%
%杨根:
%内存，操作系统
%
%英特纳雄耐尔:
%内存:32GB操作系统:windows10
% 用一段讲一下电脑配置、运行环境等。

\subsection{Evaluation metrics}
\subsubsection{AUROC} 
The AUROC metric is employed to assess image-level anomaly detection performance in this study. The AUROC value represents the area beneath the ROC curve. The ROC curve is a graphical plot with the FPR (False Positive Rate) on the horizontal axis and the TPR (True Positive Rate) on the vertical axis. Specifically, FPR measures the proportion of false positive samples among all actual negative samples, with lower values indicating better performance. Conversely, TPR quantifies the proportion of true positive samples among all actual positive samples, with higher values being preferable. For a classification model, each classification threshold corresponds to a pair of (FPR, TPR) values. By plotting all such pairs on a coordinate system and sequentially connecting the points, the ROC curve is obtained. AUROC is then calculated as the area under this ROC curve. Thus the AUROC value is threshold-invariant and reflects the intrinsic performance of the model. A higher image-level AUROC (I-AUROC) denotes a better anomaly detection performance.

\subsubsection{AUPR}The AUPR metric is used to assess pixel-level anomaly location performance in this study. The AUPR value is the area beneath the PR curve. This curve is constructed by plotting P (Precision) on the vertical axis against R (Recall) on the horizontal axis, visualizing the trade-off between precision and recall across varying thresholds. Specifically, precision represents the proportion of correctly identified positive instances among all predicted positive instances, while recall denotes the proportion of correctly identified positive instances among all actual positive instances. By calculating the pair (R, P) at different thresholds and plotting them on the coordinate, the PR curve is obtained. The AUPR value is then derived by calculating the area under this curve. Since normal regions often occupy significantly larger areas than anomalous regions in images, this leads to a class imbalance problem. AUPR emphasizes the accuracy of positive sample predictions, making it more reliable to assess model performance in this class imbalance scenario. A higher pixel-level AUPR (P-AUPR) indicates a better anomaly location performance.

%Under such circumstances, the AUPR (Area Under the Precision-Recall Curve) metric is more appropriate for evaluating model performance than other indicators.

% Notably, AUPR emphasizes the accuracy of positive sample predictions, making it more reliable to assess model performance in class imbalance scenarios. 
%with severe class imbalance—where positive samples are significantly outnumbered by negative samples. %In such cases, AUPR provides a more reliable assessment of model performance than metrics like AUROC, which may appear artificially inflated due to the dominance of negative predictions.

\subsubsection{FM} The FM (Forgetting Measure)\cite{ref3} metric evaluates both image-level and pixel-level performance degradation in this study. The FM value which measures the decline of performance on old tasks when learning new tasks, is a crucial indicator in Continual Learning. Mathematically, for the $j$-th task after the model has been trained up to task $k$: 
\begin{equation}
	\label{eqn_experiment1}
	f_j^k = \max_{l\in \{1,...,k-1\}}a_{l,j}-a_{k,j},  \forall j<k,
\end{equation}
where $a$ stands for a specific metric, i.e., the image-level AUROC or the pixel-level AUPR in this paper. Then $FM$ is calculated as follows:
\begin{equation}
	\label{eqn_experiment2}
	FM = 
	\frac{1}{k-1}\sum_{j=1}^{k-1}f_j^k.
\end{equation}
Thus, the $FM$ value indicates the extent of forgetting on prior tasks. A lower $FM$ demonstrates a better continual learning capability.

\subsection{Results and analysis}

We conducted a comparison between our method and other similar approaches. The methods are divided into three paradigms: the joint learning paradigm, the fine-tuning paradigm, and the continual learning paradigm. Under the joint learning paradigm, the model is trained on all tasks simultaneously. The performance of joint learning serves as a soft upper bound for continual learning methods, representing an idealized scenario where catastrophic forgetting is completely mitigated. In contrast, the fine-tuning paradigm trains sequentially on each task without any mechanism to retain previous knowledge, resulting in severe performance degradation on earlier tasks. Consequently, fine-tuning results typically establish a practical lower bound for continual learning evaluation. Continual learning adopts an intermediate training strategy to balance these two extremes. While it shares the same learning objective with joint learning, to master all tasks comprehensively, it operates under the same sequential training constraint as fine-tuning, processing only one task at a time. This fundamental trade-off between retaining past knowledge and acquiring new capabilities defines the core challenge of continual learning. For each paradigm, typical methods are implemented on both the MVTec AD and VisA datasets for anomaly detection. In the Joint experiment, the coreset of each task is constructed with the greedy k-center method, the random sampling method, and the proposed incremental coreset method. The total number of features for all tasks is set to 10000.  In the fine-tuning experiment, PatchCore, CFA, SimpleNet, and RD4AD are implemented in a fine-tuning paradigm. In the continual anomaly detection experiment, UCAD and the proposed CADIC are implemented. ReplayCAD, currently a state-of-the-art method in continual anomaly detection, demonstrates high-precision capabilities for both image-level anomaly detection and pixel-level anomaly localization. After sequentially learning all tasks, the final model is evaluated on the test sets across all tasks.

%Our method belongs to the continual learning paradigm. 
%该部分对比Joint,fine-tuning与持续学习下的异常检测性能。在Joint训练方式下，模型同时学习所有的任务。performance of joint learning serves as a soft upper bound for the continual learning methods. 在Fine-tuning训练方式下，模型学习完一个任务后直接再学习下一个任务，而不考虑知识遗忘的情况。Fine-tuning的结果通常为持续学习的下限。持续学习的学习方式介于两者之间。在学习目的上，他同Joint训练一样力图掌握所有任务。但在学习方式上， 他同fine-tuning一样每次只能学习一个任务。

%在Joint实验中，分别以贪婪k-center方法、随机采样方法、本文提出的incremental coreset方法一次性构建数据集上所有任务的核心集。Greedy k-center use Farthest Point Sampling technique to sample points. 随机采样方法在每个任务中选取固定数量的特征，增量核心集方法

%以随机采样法为例，针对MVTec数据集15个任务图像的特征，以随机采样法为每个任务选出667个特征，将这些特征放在一起，构成整个数据集的核心集。在检测时，根据输入特征与核心集中特征的最近距离度量异常分数，从而完成图像级的异常检测和像素级异常定位。贪婪k-center方法、本文提出的增量式coreset方法的检测流程与随机采样方法的一致，只是在构建核心集时，贪婪k-center方法是以贪婪k-center方法为每个任务筛选特征的，增量式coreset方法是以增量式coreset方法为每个任务筛选特征的。
%
%在Finetuning实验中，对PatchCore, CFA, SimpleNet, 和 RD4AD进行fine-tuning。在学习完MVTec数据集上所有任务后，检查最终得到的模型在数据集上所有任务的检测指标。The four methods includ PatchCore, CFA, SimpleNet, and RD4AD.
%
%
%在持续异常检测实验中，依次对比了DNE、UCAD以及我们提出的CADIC方法。DNE方法是早期的持续异常检测方法，但该方法只能完成图像级异常检测，不能完成像素级异常定位。UCAD方法是SOTA的持续异常检测方法，该方法通过依次学习各个任务后，可以实现对各任务的高精度图像级异常检测和像素级异常定位。
The comparison results are presented in Table \ref{tab:image MVTec} to Table \ref{tab:pixel VisA}. From these tables, we can draw the following conclusions. Firstly, the proposed method can construct a coreset incrementally. After building coresets via random sampling, greedy k-center, and our incremental coreset approach, we conduct both image-level and pixel-level anomaly detection. The results demonstrate that both greedy k-center and incremental coreset significantly outperform random sampling, validating the effectiveness of maximizing inter-point distances for feature selection. Our incremental coreset achieves comparable performance to the greedy k-center method while offering dynamic update capabilities for continual learning. In contrast, the static greedy k-center method lacks adaptability to new data streams. Secondly, the proposed method avoids Catastrophic Forgetting effectively. Direct fine-tuning of conventional anomaly detection models leads to severe performance degradation on historical tasks. After training on all 15 MVTec tasks, PatchCore, CFA, SimpleNet, and RD4AD exhibit poor retention of prior knowledge. Our method maintains near-ideal performance across all tasks, demonstrating its ability to retain discriminative features for historical categories while assimilating new data, thus effectively mitigating catastrophic forgetting. Thirdly, the proposed CADIC method achieves SOTA performance in continual anomaly detection. On the MVTec AD dataset, our CADIC achieves 0.972 image-level AUROC and 0.584 pixel-level AUROC, outperforming UCAD's 0.930 and 0.456, respectively. On VisA, we attain 0.8912 image-level AUROC (+1.72\% over UCAD) and 0.438 pixel-level AUROC (+13.8\% over UCAD). Notably, CADIC achieves 12.8\% and 13.8\% higher AUPR scores than UCAD on MVTec and VisA, respectively. These results demonstrate that, the proposed CADIC establishes new state-of-the-art benchmark of continual anomaly detection.

\begin{table*}[!t]
	\caption{Image-level metrics on MVTec AD.}
	\label{tab:image MVTec}
	\setlength{\tabcolsep}{0.6mm}
	\centering
	\begin{tabular}{l c c c c c c c c c c c c c c c c c}
		\toprule
		Methods & bottle& cable& capsule& carpet& grid& hazelnut& leather& metal\_nut& pill& screw & tile& toothbrush & transistor & wood & zipper& average & FM \\
		\midrule
		\rowcolor{gray!30}
		Joint\_PatchCore & {1.000} & 0.977 & {0.927} & {1.000} & 0.983 & {0.994} & {1.000} & {1.000} & {0.948} & {0.920} & {1.000} & 0.969 & 0.958 & {0.997} & {0.998} & {0.978} & -\\
		\rowcolor{gray!30}
		Joint\_PatchCore(R) & 0.998 & 0.936 & 0.868 & {1.000} & 0.984 & 0.979 & 1.000 & 0.993 & 0.921 & 0.653 & 0.998 & {0.975} & 0.832 & 0.995 & 0.972 & 0.940 & -\\
		\rowcolor{gray!30}
		Joint\_CADIC & {1.000} & {0.986} & 0.921 & {1.000} & {0.987} & 0.990 & {1.000} & {1.000} & 0.945 & 0.899 & {1.000} & 0.972 & {0.975} & 0.994 & 0.996 & {0.978} & -\\
		\midrule
		FT\_PatchCore & 0.163 & 0.518 & 0.350 & 0.968 & 0.700 & 0.839 & 0.625 & 0.259 & 0.459 & 0.484 & 0.776 & 0.586 & 0.341 & 0.970 & 0.991 & 0.602 & 0.383\\
		FT\_CFA  & 0.309 & 0.489 & 0.275 & 0.834 & 0.571 & 0.903 & 0.935 & 0.464 & 0.528 & 0.528 & 0.763 & 0.519 & 0.320 & 0.923 & 0.984 & 0.623 & 0.361\\
		FT\_SimpleNet  & 0.938 & 0.560 & 0.519 & 0.736 & 0.592 & 0.859 & 0.749 & 0.710 & 0.701 & 0.599 & 0.654 & 0.422 & 0.669 & 0.908 & 0.996 & 0.708 & 0.211\\
		FT\_RD4AD  & 0.401 & 0.538 & 0.475 & 0.583 & 0.558 & 0.909 & 0.596 & 0.623 & 0.479 & 0.596 & 0.715 & 0.397 & 0.385 & 0.700 & 0.987 & 0.596 & 0.393\\
		\midrule

		CFRDC\cite{CFRDC}  & 0.996 & 0.900 & 0.785 & 0.997 & 0.980 & 0.994 & 1.000 & {0.995} & {0.933} & 0.711 & 0.991 & 0.933 & \bf{0.997} & 0.982 & 0.984 & {0.945} & -\\

		IUF\cite{IUF}  & 0.909 & 0.541 & 0.520 & 0.996 & 0.695 & 0.875 & 0.997 & 0.643 & 0.547 & 0.646 & 0.940 & 0.711 & 0.660 & 0.953 & 0.795 & 0.762 & 0.067\\

		ReplayCAD\cite{ReplayCAD}  & 0.990 & 0.957 & 0.747 & 0.980 & 0.927 & 0.985 & 0.974 & 0.995 & 0.944 & 0.795 & \bf{0.999} & 0.981 & 0.957 & 0.984 & \bf{0.997} & 0.948 & 0.045\\

		DNE\cite{DNE}  & 0.990 & 0.619 & 0.609 & 0.984 & \bf{0.998} & 0.924 & 1.000 & 0.989 & 0.671 & 0.588 & 0.980 & 0.933 & 0.877 & 0.930 & 0.958 & 0.870 & 0.116\\
		UCAD\cite{UCAD}   & {1.000} &0.751 & {0.866} & 0.965 & 0.944 & {0.994} & {1.000} & 0.988 & 0.894 & 0.739 & {0.998} & \bf{1.000} & 0.874 & \bf{0.995} & 0.938 & 0.930 & \bf{0.010} \\
		DFM\cite{DFM}  & 0.997 & 0.948 & \bf{0.996} & \bf{0.999} & 0.990 & 0.977 & 1.000 & 1.000 & \bf{0.983} & 0.765 & 0.982 & 0.997 & 0.932 & 0.986 & 0.987 & 0.969 & 0.015\\
		
		CADIC(Ours) & \bf{1.000} & \bf{0.982} & 0.877 & 0.996 & 0.983 & \bf{0.994} & \bf{1.000} & \bf{1.000} & 0.942 & \bf{0.906} & 0.995 & 0.954 & \underline{0.968} & \underline{0.994} & \underline{0.990} & \bf{0.972} & \underline{0.011}\\
		\bottomrule
	\end{tabular}
\end{table*}

\begin{table*}[!t]
	\caption{Pixel-level metrics on MVTec AD.\label{tab:pixel MVTec}}
	\setlength{\tabcolsep}{0.6mm}
	\centering
	\begin{tabular}{ l c c c c c c c c c c c c c c c c c}
		\toprule
		Methods & bottle& cable& capsule& carpet& grid& hazelnut& leather& metal\_nut& pill& screw & tile& toothbrush & transistor & wood & zipper& average & FM\\
		\midrule
		\rowcolor{gray!30}			
		Joint\_PatchCore &0.820 & {0.514} & {0.525} & {0.770} & {0.300} & 0.728 & 0.224 & {0.892} & 0.811 & {0.336} & 0.620 & 0.527 & 0.637 & 0.683 & 0.531 & {0.594} & -\\
		\rowcolor{gray!30}
		Joint\_PatchCore(R) & {0.826} & 0.505 & 0.510 & 0.766 & 0.293 & 0.712 & {0.230} & 0.862 & 0.785 & 0.157 & {0.664} & {0.561} & 0.515 & 0.641 & {0.565} & 0.573 & -\\
		\rowcolor{gray!30}
		Joint\_CADIC & 0.815 & 0.510 & 0.519 & 0.754 & 0.292 & 0.744 & 0.210 & 0.886 & {0.815} & 0.307 & 0.630 & 0.530 & {0.650} & 0.675 & 0.528 & 0.591 & -\\
		\midrule	
		FT\_PatchCore  & 0.048 & 0.029 & 0.035 & 0.552 & 0.003 & 0.338 & 0.279 & 0.248 & 0.051 & 0.008 & 0.249 & 0.034 & 0.079 & 0.304 & 0.595 & 0.190 & 0.371\\
		FT\_CFA  & 0.068 & 0.056 & 0.050 & 0.271 & 0.004 & 0.341 & {0.393} & 0.255 & 0.080 & 0.015 & 0.155 & 0.053 & 0.056 & 0.281 & 0.573 & 0.177 & 0.083\\
		FT\_SimpleNet  & 0.108 & 0.045 & 0.029 & 0.018 & 0.004 & 0.029 & 0.006 & 0.227 & 0.077 & 0.004 & 0.082 & 0.046 & 0.049 & 0.037 & 0.139 & 0.060 & 0.069\\
		FT\_RD4AD  & 0.055 & 0.040 & 0.064 & 0.212 & 0.005 & 0.384 & 0.116 & 0.247 & 0.061 & 0.015 & 0.193 & 0.034 & 0.059 & 0.097 & 0.562 & 0.143 & 0.425\\
		\midrule

		CFRDC\cite{CFRDC}   & 0.737 & 0.518 & 0.425 & 0.506 & 0.243 & 0.556 & 0.372 & 0.666 & 0.417 & 0.125 & 0.454 & 0.417 & 0.710 & 0.380 & 0.390 & 0.461 & -\\
		
		IUF\cite{IUF}   & 0.289 & 0.054 & 0.040 & 0.440 & 0.084 & 0.301 & 0.330 & 0.142 & 0.048 & 0.012 & 0.310 & 0.049 & 0.065 & 0.326 & 0.08 & 0.171 & 0.059\\
		
		ReplayCAD\cite{ReplayCAD}   & 0.710 & 0.369 & 0.337 & 0.652 & \bf{0.338} & 0.635 & \bf{0.587} & 0.656 & 0.698 & \bf{0.329} & 0.531 & \bf{0.576} & 0.605 & 0.500 & \bf{0.539} & 0.537 & 0.055\\
		
		UCAD\cite{UCAD}   & 0.752 & 0.290 & 0.349 & 0.622 & 0.187 & 0.506 & 0.333 & 0.775 & 0.634 & 0.214 & 0.549 & 0.298 & 0.398 & 0.535 & 0.398 & 0.456 & \bf{0.013}\\
		
		DFM\cite{DFM}   & 0.768 & \bf{0.506} & 0.241 & \bf{0.771} & 0.228 & 0.479 & 0.432 & 0.690 & 0.576 & 0.242 & 0.623 & 0.331 & 0.501 & 0.581 & 0.511 & 0.511 & \bf{0.013}\\

		CADIC(Ours) & \bf{0.790} & 0.485 & \bf{0.506} & 0.753 & \underline{0.276} & \bf{0.749} & 0.191 & \bf{0.880} & \bf{0.810} & \underline{0.328} & \bf{0.609} & \underline{0.527} & \bf{0.650} & \bf{0.686} & \underline{0.517} & \bf{0.584} & \underline{0.015}\\	
		\bottomrule
	\end{tabular}
\end{table*}

\begin{figure}[!t]
	\centering
	\includegraphics[width=1\columnwidth]{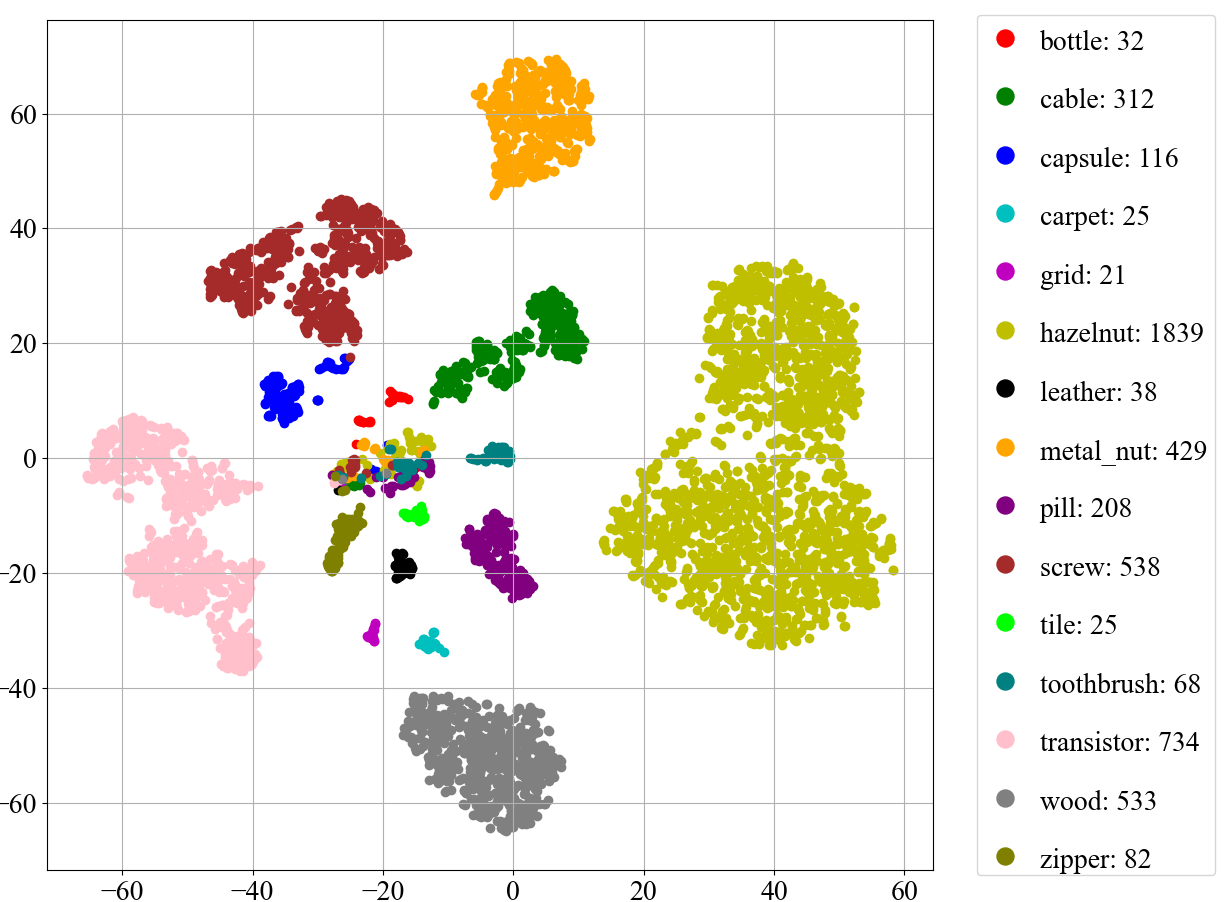}
	\caption{t-SNE visualization of features sampled from the MVTec AD dataset}
	\label{t-sne}
\end{figure}

Moreover, to analyze the distribution of different category features within the coreset constructed by the proposed method, we generate a t-SNE visualization of the coreset after completing 15 sequential tasks on the MVTec AD dataset. The results are presented in Fig. \ref{t-sne}. It shows that features of each category form distinct clusters with notable spatial separation. This segregated spatial distribution facilitates the mapping of image features to their corresponding categorical clusters, thereby enhancing detection accuracy. Moreover, categories with more diverse image content tend to have a larger number of features stored in the memory bank. The three categories with the most features, i.e., hazelnut (1,839), transistor (734), and screw (538), are all structural images with low local repetition and significant intra-class variations. These categories require extensive feature representation to comprehensively capture all normal variations. In contrast, the three categories with the fewest features, i.e., grid (21), carpet (25), and tile (25), are all texture images characterized by high local repetition and strong intra-class similarity. Consequently, only a limited number of features are required to represent all variations within these categories. Thus, the proposed method can construct a coreset with a variable number of features according to the complexity of images in different tasks, which contributes to improving the overall detection performance.

\begin{table*}[!t]
	\caption{Image-level metrics on VisA. \label{tab:image VisA}}
	\setlength{\tabcolsep}{0.9mm}
	\centering
	\begin{tabular}{ l c c c c c c c c c c c c c c c c c}
		\toprule
		Methods & candle & capsules & cashew & chewinggum & fryum & macaroni1 & macaroni2 & pcb1 & pcb2 & pcb3 & pcb4 & pipe\_fryum & average & FM\\
		\midrule
		\rowcolor{gray!30}
		Joint\_PatchCore & {0.952} & {0.878} & 0.940 & {0.983} & 0.950 & 0.902 & 0.667 & 0.963 & {0.892}& 0.894 & {0.983} & {0.995} & {0.916} & -\\
		\rowcolor{gray!30}
		Joint\_PatchCore(R) & 0.870 & 0.750 & 0.895 & 0.964 & 0.883 & 0.768 & 0.677 & 0.949 & 0.855 & 0.779 & 0.921 & 0.971 & 0.857 & -\\
		\rowcolor{gray!30}		
		Joint\_CADIC & 0.945 & 0.825 & {0.941} & 0.980 & {0.954} & {0.904} & 0.694 & {0.967} & 0.881 & 0.865 & 0.972 & 0.989 & 0.910 & -\\
		\midrule
		FT\_PatchCore & 0.401 & 0.605 & 0.624 & 0.907 & 0.334 & 0.538 & 0.437 & 0.527 & 0.597 & 0.507 & 0.588 & 0.998 & 0.589 & 0.361\\
		FT\_CFA  & 0.512 & 0.672 & 0.873 & 0.753 & 0.304 & 0.557 & 0.422 & 0.698 & 0.472 & 0.449 & 0.407 & 0.998 & 0.593 & 0.327\\
		FT\_SimpleNet  & 0.504 & 0.474 & 0.794 & 0.721 & 0.684 & 0.567 & 0.447 & 0.598 & 0.629 & 0.538 & 0.493 & 0.945 & 0.616 & 0.283\\
		FT\_RD4AD  & 0.380 & 0.385 & 0.737 & 0.539 & 0.533 & 0.607 & 0.487 & 0.437 & 0.672 & 0.343 & 0.187 & \bf{0.999} & 0.525 & 0.423\\
		\midrule	
		
		IUF\cite{IUF}  & \bf{0.994} & 0.692 & 0.758 & 0.548 & 0.677 & 0.795 & 0.606 & 0.563 & 0.766 & 0.651 & 0.512 & 0.614 & 0.681 & 0.085\\
		
		ReplayCAD\cite{ReplayCAD}  & 0.924 & 0.843 & 0.937 & 0.961 & 0.915 & \bf{0.889} & \bf{0.805} & 0.911 & 0.849 & 0.831 & \bf{0.978} & 0.991 & \bf{0.903} & 0.055\\
		
		DNE\cite{DNE}  & 0.486 & 0.413 & 0.735 & 0.585 & 0.691 & 0.584 & 0.546 & 0.633 & 0.693 & 0.642 & 0.562 & 0.747 & 0.610 & 0.179\\
		
		UCAD\cite{UCAD}  & 0.778 & \bf{0.877} & \bf{0.960} & 0.958 & \bf{0.945} & 0.823 & 0.667 & 0.905 & 0.871 & 0.813 & 0.901 & 0.988 & 0.874 & \bf{0.039}\\
		
		CADIC(ours) & 0.859 & 0.817 & 0.926 & \bf{0.972} & 0.910 & \underline{0.839} & {0.733} & \bf{0.946} & \bf{0.878} & \bf{0.869} & \underline{0.952} & \bf{0.993} & \underline{0.891} & \underline{0.043}\\		
		\bottomrule
	\end{tabular}
\end{table*}
\begin{table*}[!t]
	\caption{Pixel-level metrics on VisA. \label{tab:pixel VisA}}
	\setlength{\tabcolsep}{0.9mm}
	\centering
	\begin{tabular}{ l c c c c c c c c c c c c c c c c c}
		\toprule
		Methods & candle & capsules & cashew & chewinggum & fryum & macaroni1 & macaroni2 & pcb1 & pcb2 & pcb3 & pcb4 & pipe\_fryum & average & FM\\
		\midrule
		\rowcolor{gray!30}
		Joint\_PatchCore & 0.206 & 0.617 & 0.717 & 0.768 & 0.485 & {0.056} & 0.015 & 0.790 & 0.084 & 0.561 & {0.359} & 0.620 & 0.440 & - \\
		\rowcolor{gray!30}
		Joint\_PatchCore(R) & {0.213} & 0.581 & 0.685 & 0.770 & 0.476 & 0.029 & 0.010 & 0.770 & 0.070 & 0.496 & 0.293 & {0.637} & 0.419 & - \\
		\rowcolor{gray!30}		
		Joint\_CADIC & 0.196 & 0.618 & 0.716 & {0.771} & {0.490} & 0.054 & 0.015 & 0.793 &{ 0.088} & {0.573} & 0.349 & 0.610 & {0.439} & - \\
		\midrule
		FT\_PatchCore  & 0.012 & 0.007 & 0.055 & 0.315 & 0.082 & 0.000 & 0.000 & 0.008 & 0.004 & 0.007 & 0.010 & 0.585 & 0.090 & 0.311\\
		FT\_CFA  & 0.017 & 0.005 & 0.059 & 0.243 & 0.085 & 0.001 & 0.001 & 0.013 & 0.006 & 0.008 & 0.015 & 0.592 & 0.087 & 0.184\\
		FT\_SimpleNet  & 0.001 & 0.004 & 0.017 & 0.007 & 0.047 & 0.000 & 0.000 & 0.013 & 0.003 & 0.004 & 0.009 & 0.058 & 0.014 & 0.016\\
		FT\_RD4AD  & 0.002 & 0.005 & 0.061 & 0.045 & 0.098 & 0.001 & 0.001 & 0.013 & 0.008 & 0.008 & 0.013 & 0.576 & 0.069 & 0.201\\		
		\midrule	

		IUF \cite{IUF}  & 0.012 & 0.017 & 0.043 & 0.033 & 0.107 & 0.011 & 0.004 & 0.019 & 0.009 & 0.018 & 0.021 & 0.117 & 0.034 & \bf{0.003}\\
		
		ReplayCAD\cite{ReplayCAD}  & \bf{0.241} & 0.430 & 0.555 & 0.674 & 0.462 & \bf{0.178} & \bf{0.099} & 0.793 & \bf{0.199} & 0.422 & 0.303 & 0.625 & 0.415 & 0.050\\
		
		UCAD\cite{UCAD}  & 0.067 & 0.437 & 0.580 & 0.503 & 0.334 & 0.013 & 0.003 & 0.702 & {0.136} & 0.266 & 0.106 & 0.457 & 0.300 & 0.015\\
		
		CADIC(ours) & \underline{0.193} & \bf{0.631} & \bf{0.718} & \bf{0.763} & \bf{0.476} & \underline{0.047} & \underline{0.019} & \bf{0.794} & 0.087 & \bf{0.559} & \bf{0.337} & \bf{0.631} & \bf{0.438} & \underline{0.014}\\		
		\bottomrule
	\end{tabular}
\end{table*}

\begin{figure}[!t]
	\centering
	\includegraphics[width=1\columnwidth]{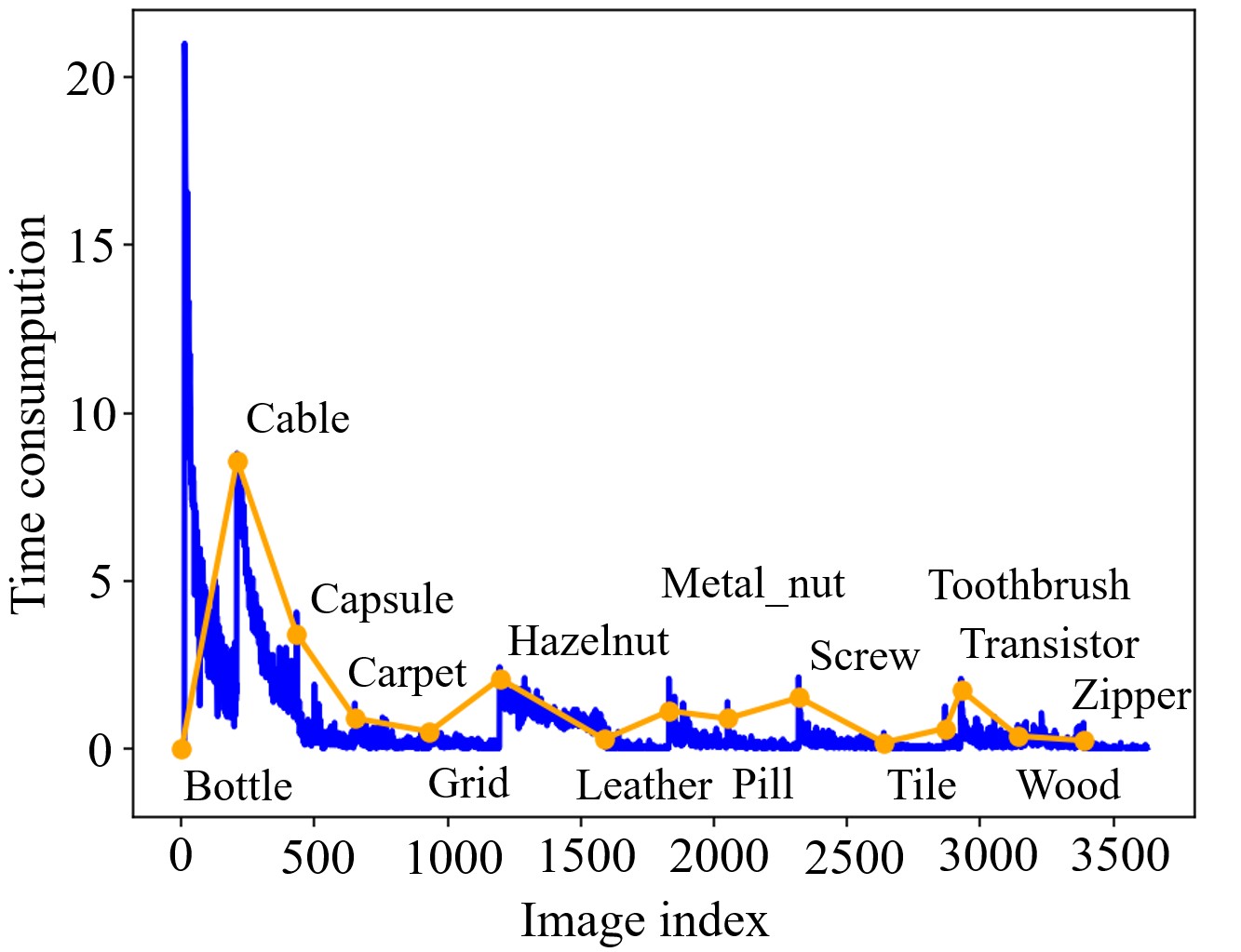}
	\caption{Time consumption curve for every 100 images sampled. The horizontal coordinate axis is the image batch number (each batch corresponds to 100 images), and the vertical coordinate axis is the time consumed (seconds).}
	\label{time_consumption}
\end{figure}

Furthermore, to analyze the time consumption of the proposed method, we plot the time consumption curve as shown in Fig. \ref{time_consumption}. The x-axis represents the image sequence number, and the y-axis represents the time consumption. The blue curve illustrates the time consumption of each single image, and the orange points denote the start time of each category. From the figure, we can conclude the following results. In general, learning individual images from earlier tasks requires more time compared to those from later tasks. As shown in the figure, the earlier tasks like bottle, cable, and capsule take longer than later tasks such as toothbrush, transistor, and zipper. This phenomenon arises because during early tasks, the features in the coreset are too close to each other, leading to frequent updates. While in the later tasks, the distances between adjacent features in the coreset are large enough, leading to fewer element updates. In detail, the processing time of each category decreases rapidly. It shows that, for any specific category, the processing times for the initial few images are significantly longer, but they decrease rapidly afterward. This occurs because at the beginning of learning each category, the coreset contains few representative features of the category. As a result, a substantial amount of features from images of this category are incorporated into the coreset. However, as the features of this category gradually increase, it becomes increasingly difficult for features of this category to be incorporated into the coreset. The reason why the time consumption for the first few images of ``Bottle" is close to zero is that the features of these images are directly placed into the coreset for  initialization. Moreover, learning complex images takes longer than learning simple ones. As can be seen from the figure, the learning time of complex images such as hazelnut, screw, and transistor is longer than that of simpler images like grid, carpet, and tile. This is because the more complex an image is, the greater the differences among its features, necessitating to preserving more features in the coreset. Conversely, the simpler an image is, the smaller the differences among its features, with only a few highly representative features being likely to enter the coreset.

\begin{figure*}[!t]
	\centering
	\includegraphics[width=0.95\textwidth]{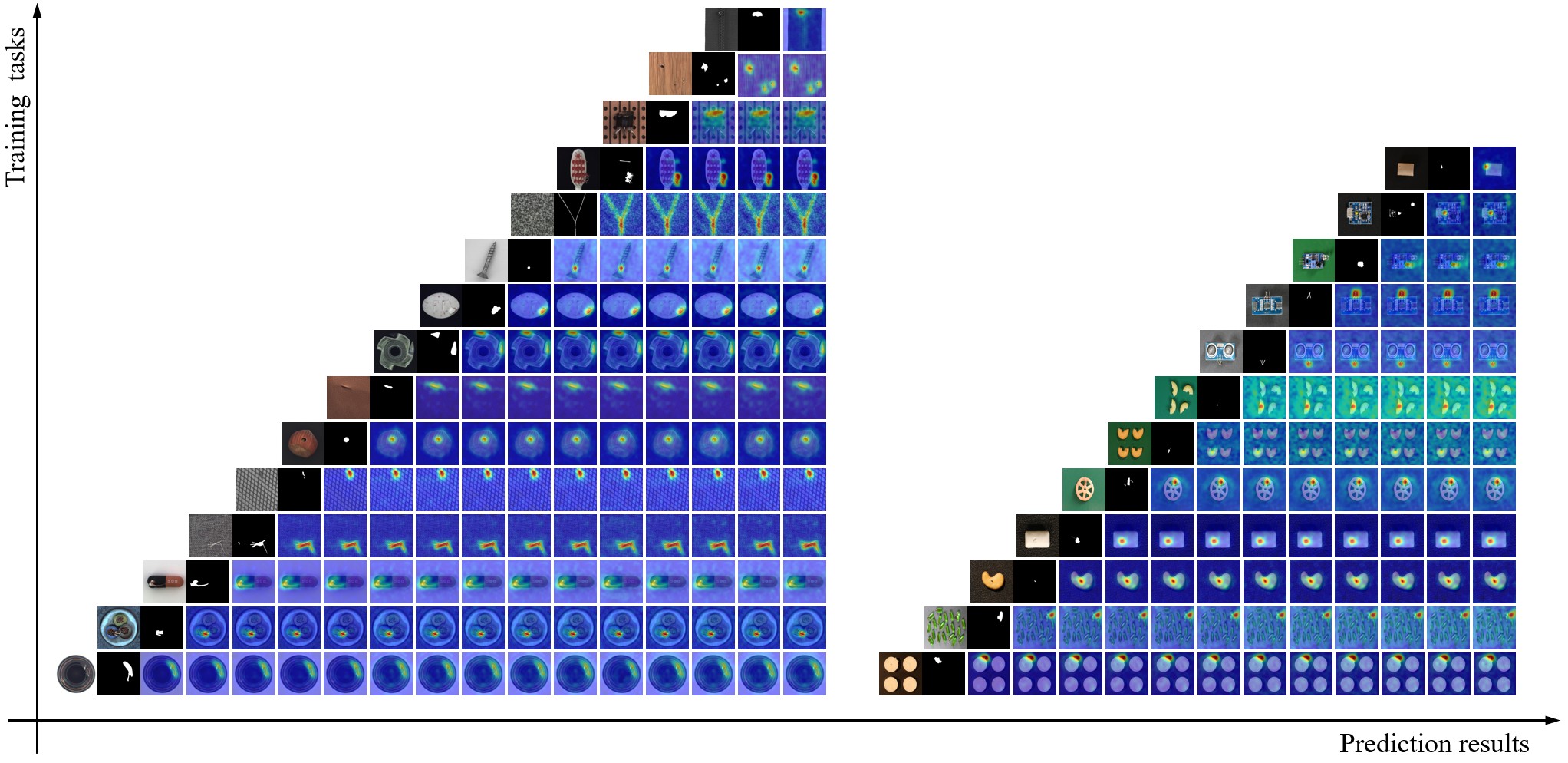}
	\caption{The results of anomaly localization of CADIC in MVTec AD and VisA dataset, respectively. }
	\label{results-of-detection}
\end{figure*}

Fig. \ref{results-of-detection} illustrates the anomaly localization results of our method on learned tasks after completing the training of a new task on the MVTec AD and VisA datasets. We can observe the following two key points. Firstly, the localization results on old tasks remain largely unchanged even after learning multiple new tasks. These results indicate that the coreset constructed by CADIC effectively preserves the most critical historical knowledge. Secondly, CADIC demonstrates excellent localization capability for subtle defects, reflecting its robustness in perceiving fine-grained features.

\subsection{Ablation Study}
In this section, we conduct comprehensive ablation experiments on MVTec AD and VisA benchmarks. 

\subsubsection{Extractor}
We perform ablation experiments to explore the influence of different embedding extractors on the continual learning ability of our method. \textit{ResNet50} has been proven by numerous prior works to be an effective feature extractor. By concatenating feature maps of the second and the third block, the embeddings are obtained. \textit{ViT-base-patch16} and \textit{ViT-base-patch8} are both transformer-based neural networks. For \textit{ViT-base-patch16}, each embedding in the feature map corresponds to a $16\times16$ patch of the original image, resulting in coarser representations; while for \textit{ViT-base-patch8}, each embedding corresponds to an $8\times8$ patch, yielding finer representations. Table \ref{tab:extractor} shows the performance variations across these three extractors. It can be observed that the \textit{ViT-base-patch8} achieves optimal performance in both anomaly detection and localization tasks. Thus, we choose \textit{ViT-base-patch8} as the embedding extractor.

%ViT-base-patch16与ViT-base-patch8均是基于transformer的神经网络。他们之间的不同之处在于，前者特征图上每个embedding对应于原图上16×16的图像块，更粗糙；而后者特征图上每个embedding对应于原图上8×8的图像块，更细致。我们以这两个网络的第9层提取图像特征。

%  It can be observed that the \textit{ViT-base-patch8} achieves optimal performance in both anomaly detection and localization tasks.
% attributed to its finer-grained 8x8 image patch partitioning and effective self-attention mechanism.

%其中，对于WideResNet50，我们拼接第二个网络块与第三个网络块的图像特征图以获得最终的特征嵌入，这也是被其他研究验证过的方法。对于Vit16和vit8，我们均以其第9层网络提取图像特征。所有网络提取的图像特征数量均为10000个。
%
%\begin{table*}[!t]
%	\caption{Ablation study for extractor on MVTech AD. \label{tab:extractor MVTec AD}}
%	\setlength{\tabcolsep}{1.5mm}
%	\centering
%	\begin{tabular}{c c c c c c c}
%		\hline
%		Extractor &  I-AUROC & I-FM &  P-AUPR & P-FM & Training Time (seconds) & Inference Speed (FPS)\\
%		\hline
%		CADIC-wr50 & 0.949 & 0.033 & 0.498 & 0.043 & 2,337 & 6.5\\
%		CADIC-vb16 & 0.940 & 0.026& 0.493 & \bf{0.010} & \bf{1,859} & \bf{7.8}\\
%		CADIC-vb8 & \bf{0.972} & \bf{0.011} & \bf{0.584} & 0.015 & 2,503 & 6.6\\
%		\hline
%	\end{tabular}
%\end{table*} 
%\begin{table*}[!t]
%	\caption{Ablation study for extractor on VisA. \label{tab:extractor VISA}}
%	\setlength{\tabcolsep}{1.5mm}
%	\centering
%	\begin{tabular}{c c c c c c c}
%		\hline
%		Extractor &  I-AUROC & I-FM &  P-AUPR & P-FM & Training Time (seconds) & Inference Speed (FPS)\\
%		\hline
%		CADIC-wr50 & 0.831 & 0.097 & 0.370 & 0.031 & 3,003 & 7.1\\
%		CADIC-vb16 & 0.844 & 0.048 & 0.350 & 0.062 & \bf{2,723} & \bf{8.0}\\
%		CADIC-vb8 & \bf{0.891} & \bf{0.043} & \bf{0.438} & \bf{0.014} & 3,165 & 6.5\\
%		\hline
%	\end{tabular}
%\end{table*}

\begin{table}[]
	\centering
	\caption{ Performance Comparison of Different Extractors}
	\label{tab:extractor}
	\begin{tabular}{l c c c c }
		\toprule
		\multirow{2}{*}{Extractor}  & \multicolumn{2}{c}{\textbf{MVTec AD}} & \multicolumn{2}{c}{\textbf{VISA}}  \\ \cmidrule
		(lr){2-3} \cmidrule(lr){4-5}
		
		&  I-AUROC &   P-AUPR  &  I-AUROC &   P-AUPR  \\ 
		\midrule		
		WideResNet50 & 0.949 & 0.498  & 0.831  & 0.370 \\
		ViT-base-patch16 & 0.940 & 0.493  & 0.844  & 0.350 \\
		ViT-base-patch8 & \bf{0.972} & \bf{0.584} & \bf{0.891} & \bf{0.438}\\		
		\bottomrule
	\end{tabular}
\end{table}

\begin{table}[]
	\centering
	\caption{Performance Comparison of Different layers}
	\label{tab:layer}
		\setlength{\tabcolsep}{10pt}
	\begin{tabular}{c c c c c }
		\toprule
		\multirow{2}{*}{Layer}  & \multicolumn{2}{c}{\textbf{MVTec AD}} & \multicolumn{2}{c}{\textbf{VISA}}   \\ \cmidrule
		(lr){2-3} \cmidrule(lr){4-5}		
		&  I-AUROC &   P-AUPR  &  I-AUROC &   P-AUPR  \\ 
		\midrule		
		1 & 0.836  & 0.515  & 0.722  & 0.261 \\
		3 & 0.904  & 0.591  & 0.799  & 0.363 \\
		5 & 0.932 & 0.587  & 0.831  & 0.391 \\
		7 & 0.947  & \bf{0.612}  & 0.881 & \bf{0.446} \\
		9 & \bf{0.972} & 0.584 &   \bf{0.891} &  0.438 \\
		11 & 0.946  & 0.535  & 0.870 &  0.404 \\
		\bottomrule
	\end{tabular}
\end{table}

\begin{table*}[]
	\centering
	\caption{Performance Comparison of Different coreset size}
	\label{tab:memorysize}
	\begin{tabular}{c c c p{1.8cm}<{\centering} p{1.6cm}<{\centering} c c p{1.8cm}<{\centering} p{1.6cm}<{\centering} c c c c }
		\toprule
		\multirow{3}{*}{ Coreset size}  & \multicolumn{4}{c}{\textbf{MVTec AD}} & \multicolumn{4}{c}{\textbf{VISA}}  \\ \cmidrule
		(lr){2-5} \cmidrule(lr){6-9}		
		&  I-AUROC &   P-AUPR & Training Time (seconds) & Inference Speed (FPS) &  I-AUROC &   P-AUPR & Training Time (seconds) & Inference Speed (FPS) \\ 
		\midrule		
		2500 &0.955 &  0.559 &  \bf{63} & \bf{7.9} & 0.828  & 0.415  & \bf{79} & \bf{6.8}\\
		5000 & 0.969  & 0.577  & 379 & 7.8 & 0.856 & 0.427  & 474 & 6.4\\
		10000 & 0.972 & 0.584  & 2,503 & 6.6 & 0.891  & 0.438  & 3,165 & 6.5\\
		20000 & \bf{0.978}  & \bf{0.593} & 17,038 & 6.4 & \bf{0.916} & \bf{0.439}  & 22,254 & 6.5\\
		\bottomrule
	\end{tabular}
\end{table*}

%\begin{table}[h]
%	\centering
%	\begin{tabularx}{\textwidth}{4|2|X} % X表示自动宽度，|为分隔线样式
%		1 & 2 & 3 \\
%		1,1 & 1,2 & 1,3 \\
%		2,1 & 2,2 & 2,3 \\
%	\end{tabularx}
%	\caption{c}
%\end{table}

%\begin{table}[!h]
%	\centering
%	\setlength{\tabcolsep}{10pt} % 减小列间距 (默认为 6pt)
%	\begin{tabular}{cc}
%		\hline
%		{\textbf{System}} & {\textbf{TeX}} \\
%		\hline
%		All & TeX Live \\
%		macOS & MacTeX \\
%		Windows & MikTeX \\
%		\hline
%	\end{tabular}
%\end{table}

\subsubsection{Layer}
We also examined the effectiveness of different layers of \textit{ViT-base-patch8}. The network comprises 12 feature extraction layers, each capable of extracting feature maps with dimensions of $28\times28\times768$. For simplicity, we compare only the performance of odd-numbered layers. The results are shown in the Table\ref{tab:layer}. It shows that as the layers get deeper, the extracted features become increasingly effective in our method. However, when layers are too close to the output layer, the performance declines. Specifically, the 7th layer achieves the best pixel-level metrics, while the 9th layer excels in image-level metrics. Compared to pixel-level anomaly localization, we prioritize image-level anomaly detection. Therefore, we select the 9th layer as the feature extraction layer in this study.

%该网络一共有12个特征提取层，每层都能提取到尺寸为28×28×768的特征图。为简化起见，我们仅比较了各奇数层的性能，结果如表所示。从表中数据可以看出，随着特征网络层数的加深，网络提取的特征越有效。但当网络层数过于接近输出层时，网络提取的特征性能下降。其中，第7层具有最好的像素级指标，第9层具有最好的图像级指标。相比于像素级的异常定位，我们更注重图像级的异常检测。故本文选择第9层作为特征提取层。

%和第9层的异常检测指标优于其他层。

%在基于embedding的异常检测方法中，更适合采用transformer网络中间偏后的网络提取图像特征。，中间偏后的层网络提取的图像特征更适合, which substantiates the hypothesis proposed in former research that intermediate-layer features demonstrate optimal adaptability for complex visual tasks.

%The experimental comparison results are shown in Table \ref{tab:layer}. It can be seen that the different layers of \textit{ViT-base-patch8} exert significant impacts on model performance. In anomaly localization tasks, the 9th layer achieves optimal performance with image-AUROC scores of 97.2\% and 89.1\% on MVTec and Visa datasets, respectively. In addition, the 7th layer attains the highest p-AUPR in anomaly localization, but it exhibits higher p-FM values compared to the 9th layer. These results indicate that intermediate-layer features effectively integrate spatial details from shallow layers with semantic abstractions from deeper layers, which substantiates the hypothesis proposed in former research that intermediate-layer features demonstrate optimal adaptability for complex visual tasks. Notably, as depth increases, metrics also get smaller, suggesting that deeper features reduce reliance on non-critical features of historical tasks through more intensive attention interactions.

\subsubsection{ Coreset size }
Finally, we study the impact of memory bank size on the model's continual learning capability. Table \ref{tab:memorysize} shows there is a significant positive correlation between coreset size and model performance. When the coreset size is small, the model trains faster but exhibits lower anomaly detection metrics. At a coreset size of 20,000, the model achieves the best anomaly detection performance, but its training time becomes excessively long. When the coreset size is 10,000, the model achieves both relatively high detection metrics and acceptable training time. Therefore,  this paper sets the coreset size to 10,000.

%随着coreset size的增大，模型的性能提升。当coreset size较小时，模型的训练速度很快，但异常检测指标较低。当coreset size为20000时，模型具有最好的异常检测性能，但其花费的训练时间过长。综合考虑训练速度和检测精度，本文将coreset size设置为10,000.
%当coresetsize为10000时，模型既具有较高的检测指标，有具有接受范围内的训练时间。因此，综合考虑训练速度和检测精度，本文将coreset size设置为10,000.

%Concurrently, the FM metrics also show a monotonically decreasing trend, which indicates that when the model enhances the learning ability of new tasks, it can also maintain the knowledge representation of historical tasks more effectively. However, these performance gains are accompanied by nonlinearly escalating computational costs-training time surges from 63s to 17,038s. Considering both detection performance and training time, the coreset size is set to 10,000.

\subsection{ Further study }
%为了研究本文提出的方法在超长期异常检测任务上的性能，我们在MVTec 与 VISA组成的混合数据集上进行了持续异常检测实验，结果如表所示。从表中数据可以看出，本文方法持续学习27个任务后，图像级AUROC的平均值和像素级AUPR的平均值分别为0.921和0.492，依然具有很高的值。该实验表明，本文方法在超长异常检测任务上具有优异的检测效果。
To investigate CADIC's performance on ultra-long-term anomaly detection tasks, we further conducted continual anomaly detection experiments on a hybrid dataset composed of MVTec and VISA. The results are shown in the table \ref{tab:Ultra-long-term}. It shows that, after continual learning across all 27 tasks, CADIC still achieves competitive metrics, with an average image-level AUROC of 0.921 and an average pixel-level AUPR of 0.492. This experiment demonstrates that the proposed CADIC exhibits excellent detection performance in ultra-long-term anomaly detection tasks.

%To explore the performance limits of our method, we conducted an ultra-long continual learning experiment on a combined dataset consisting of MVTec AD and VisA. Considering inference speed, we employed an independent memory bank during the inference phase. As shown in Table \ref{tab:10}, our methods achieve an impressive I-AUROC of 96.25\% and a P-AP of 51.9\%. These results are approximately equal to the average (I-AUROC: 96.45, P-AP: 52.4) obtained from performing two separate long-term continual learning experiments on MVTec AD and Visa, which demonstrates EmbeddingAD has a stable memory ability for ultra-long incremental training.
\begin{table*}
	\centering
	\caption{Ultra-long-term continual anomaly detection performance}
	\label{tab:Ultra-long-term}
	\begin{tabular}{lccccccccccc}
		\toprule
		Category &   bottle   & cable &  capsule   &   carpet   & grid & hazelnut & leather & metal nut &    pill    & - & - \\ 
		\midrule
		I-AUROC & 0.998 & 0.984 & 0.887 & 0.995 & 0.967 & 0.999 & 1.000 & 1.000 & 0.957 & -  & - \\
		P-AUPR & 0.795 & 0.502 & 0.476 & 0.760 & 0.224 & 0.710 & 0.136 & 0.834 & 0.803 & - & - \\ 
		\bottomrule
		Category &   screw    & tile  & toothbrush & transistor & wood &  zipper  & candle  & capsules  &   cashew   &    -  & -   \\ 
		\midrule
		I-AUROC & 0.901 & 0.995 & 0.908 & 0.949 & 0.990 & 0.968 & 0.852 & 0.846 & 0.871 & -  & -\\
		P-AUPR & 0.277 & 0.543 & 0.508 & 0.643 & 0.650 & 0.403 & 0.203 & 0.575 & 0.766 & - & - \\ 
		\bottomrule
		Category & chewinggum & fryum & macaroni1  & macaroni2  & pcb1 &   pcb2   &  pcb3   &   pcb4    & pipe\_fryum & average & FM \\ 
		\midrule
		I-AUROC & 0.993 & 0.902 & 0.764 & 0.702 & 0.922 & 0.809 & 0.827 & 0.909 & 0.989 & 0.921  & 0.023 \\
		P-AUPR & 0.721 & 0.472 & 0.026 & 0.009 & 0.794 & 0.081 & 0.436 & 0.305 & 0.644 & 0.492  & 0.011 \\  
		\bottomrule
	\end{tabular}
\end{table*}
\begin{figure}[!t]
	\centering
	\includegraphics[width=1\linewidth]{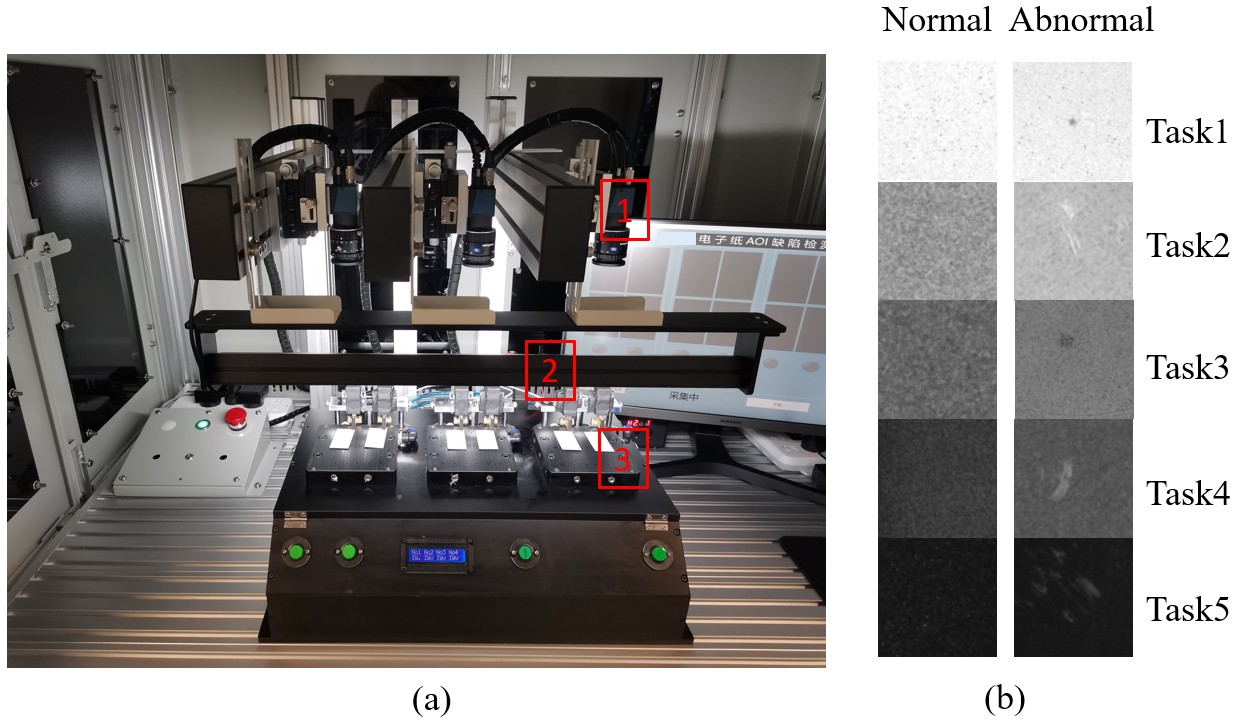}
	\caption{(a) Automatic optical inspection platform. 1: Camera, 2: Light source, 3: The inspected E-paper. (b) Five tasks from the E-paper dataset}
	\label{Epaperresults}
\end{figure}

\subsection{ Application  }
\begin{figure*}[!t]
	\centering
	\includegraphics[width=0.8\textwidth]{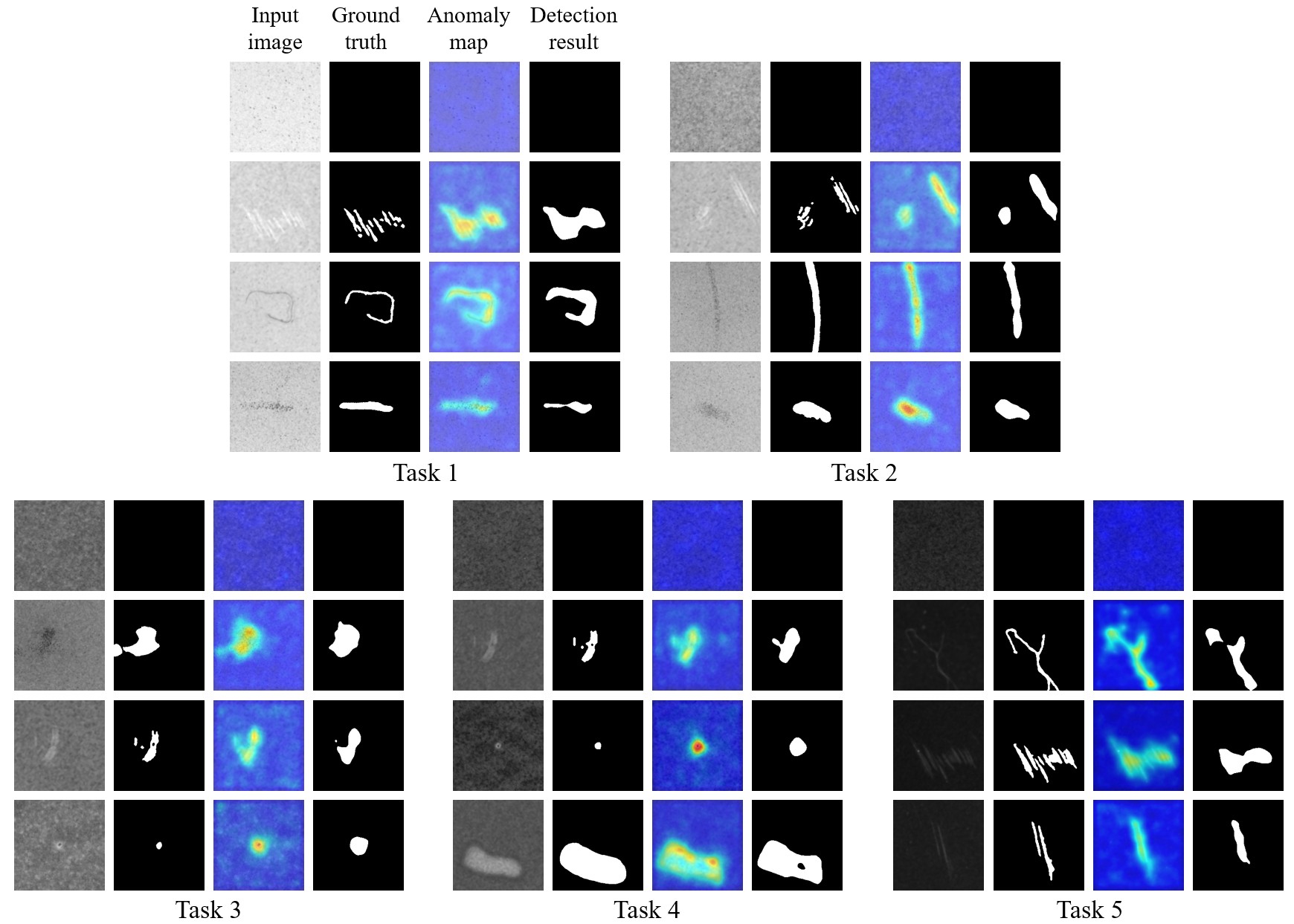}
	\caption{The results of anomaly localization of CADIC in E-paper dataset.}
	\label{results}
\end{figure*}

To further demonstrate the practical value of CADIC, we apply CADIC to detect the anomalies of real-world Electronic Paper (E-paper). Fig. \ref{Epaperresults}(a) shows the automatic optical  inspection platform. Based on this platform, we collect a large number of electronic paper images and construct an electronic paper anomaly detection dataset. The dataset includes $750$ normal images and $250$ abnormal images. For E-paper, different defects are only visible under specific grayscale. Therefore, we partition the dataset into five independent tasks according to the grayscale levels of the images. Each task contains 100 normal images for training, $50$ normal images and $50$ abnormal images for inference. Fig. \ref{Epaperresults}(b) illustrates typical images of each task. We use CADIC to learn the five tasks sequentially, and the detection results are shown in table \ref{Epaper}. We observe that CADIC shows outstanding performance across all five tasks. Notably, almost all the tasks' image-level AUROC are $1.0$. The results indicate that CADIC achieves perfect classification performance in image-level anomaly discrimination. Moreover, Fig. \ref{results} shows the localization result of the five tasks. we can see that CADIC can  localize various types of defects across different tasks effectively. This experiment demonstrates that CADIC exhibits strong practicality for industrial application.

\begin{table*}[!t]
	\caption{Performace of the proposed CADIC in the E-paper datasets}
	\setlength{\tabcolsep}{2.0mm}
	\centering
	\begin{tabular}{ c c c c c c c c}
		\hline
		Measurements & task1  & task2 &  task3 & task4 & task5 & average & FM\\
		\hline
		I-AUROC & 1.000 & 1.000 & 1.000 & 1.000 & 0.999 & 1.000 & 0\\
		P-AUPR & 0.490 & 0.555 & 0.596 & 0.592 & 0.371 & 0.521 & 0.010 \\		
		\hline
	\end{tabular}
	\label{Epaper}
\end{table*}

%\subsection{ Application  }
%Anomaly detection of E-paper.

%\begin{table}[!t]
%	\caption{Ablation study for ViT encoder layer on MVTech AD.\label{tab:layer VISA}}
%	\setlength{\tabcolsep}{2.0mm}
%	\centering
%	\begin{tabular}{ c c c c c c c c}
%		\hline
%		Tasks & Task 1  & Task 2 &  Task 3 & Task 4 & Task 5 & average & FM\\
%		\hline
%		I-AUROC & 1.000 & 1.000 & 1.000 & 1.000 & 0.999 & 1.000 & 0\\
%		P-AUPR & 0.490 & 0.555 & 0.596 & 0.592 & 0.371 & 0.521 & 0.010 \\		
%		\hline
%	\end{tabular}
%\end{table}

\section{conclusion}
In this study, we present a novel continual anomaly detection method termed CADIC, which learns sequential tasks through incremental memory bank updates. By employing the pre-trained high-resolution ViT-Base-Patch8-224 backbone as the feature extractor, we achieve highly refined embeddings with enhanced contextual awareness. We introduce a mini-batch updating mechanism that constructs memory banks of multiple tasks incrementally, enabling effective anomaly detection in a continual learning setting. Experiments are conducted on both the MVTec AD dataset and the VisA dataset. The results show that the proposed method surpasses existing SOTA continual anomaly detection methods at both image-level and pixel-level. Our method is intuitive, easy to implement, and highly accurate, demonstrating strong practical applicability.

In future work, we aim to further optimize the memory bank update mechanism. The proposed method involves iteratively computing the Euclidean distances between features in the current update batch and those stored in the memory bank. The overall computation costs is high for large-scale memory banks. Therefore, we will investigate more efficient updating strategies to accelerate the memory bank updating processes.

%\begin{table*}[!t]
%	\caption{Pixel-level AUPR and corresponding avg FM on MVTec AD dataset\label{appendix-table1}}
%	\setlength{\tabcolsep}{0.6mm}
%	\centering
%	\begin{tabular}{c c c c c c c c c c c c c c c c c c}
%		\hline
%		Extractor & Bot.& cab.& cap.& car.& gri.& haz.& lea.& met.& pil.& scr. & til. & too. & tra. & woo. & zip. & average & avg. FM\\
%		
%		WideResNet50 & 0.971 & \bf{0.988} & 0.660 & 0.944 & 0.920 & \bf{1.000} & 0.987 & 0.984 & 0.793 & 0.713 & 0.984 & 0.940 & \bf{0.995} & 0.987 & 0.919 & 0.919 & 0.054 \\
%		
%		Vit-base-patch16-224 & 0.957 & 0.865 & 0.567 & 0.909 & 0.767 & 0.995 & 0.991 & 0.891 & 0.717 & 0.601 & 0.987 & 0.811 & 0.834 & 0.938 & 0.943 & 0.851 & 0.08 \\
%		
%		Vit-base-patch8-224 & \bf{0.996} & 0.950 & \bf{0.824} & \bf{0.990} & \bf{0.985} & 0.993 & \bf{0.999} & \bf{1.000} & \bf{0.922} & \bf{0.882} & \bf{0.994} & \bf{0.956} & 0.973 & \bf{0.995} & \bf{0.987} & \bf{0.963} & \bf{0.018} \\
%		\hline
%	\end{tabular}
%\end{table*}

\end{document}